\documentclass[10pt]{article}

\usepackage{style/basic}

\definecolor{family}{HTML}{ECECEC}
\definecolor{pajamablue}{HTML}{E8F0FE}

\hypersetup{
    colorlinks=true,
    linkcolor=purple,
    urlcolor=teal,
    linktoc=all,
    citecolor=teal
}

\newtheorem*{theorem*}{Theorem}

\def \b0 {{\bf 0}}

\makeatletter

\newcommand{\changeoperator}[1]{%
  \csletcs{#1@saved}{#1@}%
  \csdef{#1@}{\changed@operator{#1}}%
}

\newcommand{\changed@operator}[1]{%
  \mathop{%
    \mathchoice{\textstyle\csuse{#1@saved}}
               {\csuse{#1@saved}}
               {\csuse{#1@saved}}
               {\csuse{#1@saved}}%
  }%
}

\makeatother

\newcommand\blfootnote[1]{%
  \begingroup
  \renewcommand\thefootnote{}\footnote{#1}%
  \addtocounter{footnote}{-1}%
  \endgroup
}

\newcommand{\EqualContribution}{\textsuperscript{*}Equal Contribution. Corresponding Authors: <thuang273, sqiu53>@wisc.edu}

\let\oldnl\nl
\newcommand{\nonl}{\renewcommand{\nl}{\let\nl\oldnl}}

\newif\ifsinglecolumn
\singlecolumntrue

\title{Codifying the Judge: Scalable Evaluation via Program Distillation}

\author[$*\dagger$]{Tzu-Heng~Huang}
\author[$*\dagger$]{Shengqi~Qiu}
\author[$\dagger$]{Frederic~Sala}

\affil[$\dagger$]{University of Wisconsin-Madison}

\begin{document}
	\maketitle
	\blfootnote{\EqualContribution}

\begin{abstract}
LLM-as-a-judge has become the standard for automated evaluation, but it suffers from high cost, inference latency, and opaque decisions---limitations that undermine its scalability and reliability.
We address these with a simple, efficient alternative: program distillation.
Instead of prompting an LLM at evaluation time, we distill its decision logic into a committee of  programs that can score candidates directly.
These programmatic judges offer transparency, are easily inspected or edited, and eliminate per-sample API costs.
Building on this notion, we introduce PAJAMA, a system that synthesizes programs as judges, aggregates their decisions into a joint verdict, and incorporates a fallback mechanism to selectively escalate low-confidence cases to an LLM.
Across five datasets and eight model families, we show that programmatic judges match the performance of a 13B-size LLM judge at 47$\times$ higher throughput.
When using program outputs as routing signals, PAJAMA improves both accuracy and throughput and advances the Pareto frontier.
Beyond evaluation, programmatic judges produce cheap and effective reward signals: on RewardBench, a reward model distilled from programs' verdicts outperforms one trained on a proprietary LLM's labels at two orders of magnitude lower API cost.
\end{abstract}
\footnotetext[1]{
Our source code is available \href{https://github.com/SprocketLab/PAJAMA}{here}.
Project page and our demonstration can be found in~\url{https://sprocketlab.github.io/PAJAMA/}.}

\section{Introduction}
\label{sec:introduction}

The LLM-as-a-judge paradigm is a core building block of modern machine learning pipelines~\citep{gu2024survey, li2024llms}.
LLM judges assign pairwise preference labels~\citep{zhu2025judgelm, kim2024prometheus, wang2024pandalm}, supply supervision signals for reward model distillation~\citep{ye2025learning, wu-etal-2025-meta, wang2024self}, and score trajectories against rubric criteria to provide feedback in reinforcement learning~\citep{gunjal2026rubrics, huang2026rubicap, shao2025dr}.
As evaluation moves into online settings and iterative pipelines grow ever more prominent, however, the fundamental drawbacks of LLM judges are becoming difficult to ignore.

We identify four key challenges that limit the \textbf{\emph{scalability}} and \textbf{\emph{reliability}} of LLM judges:
\begin{itemize}[nosep,topsep=0pt,leftmargin=*]
    \item \textbf{Inference Cost.}
    When using proprietary frontier models such as OpenAI GPT-6 Astra, Anthropic Claude Fable 5 in the loop, scaling evaluation to millions of samples yields prohibitive API spending~\citep{salinas2025tuning}.
    Open-weight deployments avoid this API bill but still impose substantial inference latency and GPU requirements.
    \item \textbf{Opaque Logic.}
    While LLMs produce justifications, their internal decision process is opaque.
    It is hard to verify whether a verdict relies on the stated rationale or is a product of hallucination~\citep{zhao2025one, maloyan2025investigating}.
    %
    %
    \item \textbf{Systemic Bias.}
    LLM judges are sensitive to stylistic biases, favoring verbosity, rich formatting, or emotionally charged language---all of which undermine reliability~\citep{chen-etal-2024-humans, ye2025justice, shi-etal-2025-judging, schroeder2024can, zhao2026care}.
    \item \textbf{Re-inference Tax.}
    Existing prompting pipelines are inflexible.
    Revising a single rubric criterion requires re-running inference over the entire dataset, incurring redundant costs and wasted cycles.
\end{itemize}

We address these obstacles by \textbf{\emph{shifting from model-based inference to synthesized program execution}}.
Instead of asking an LLM to independently assess each candidate, we ask it to generate the judging logic it would apply, and convert that logic into an executable program.
In other words, the LLM is now queried once at synthesis time, turning repeated model inference into a one-time investment.
Afterwards, we can invoke programs locally to produce verdicts on every candidate, producing judgments without further model calls.

\begin{figure*}[t!]
    \centering
    \includegraphics[width=\linewidth]{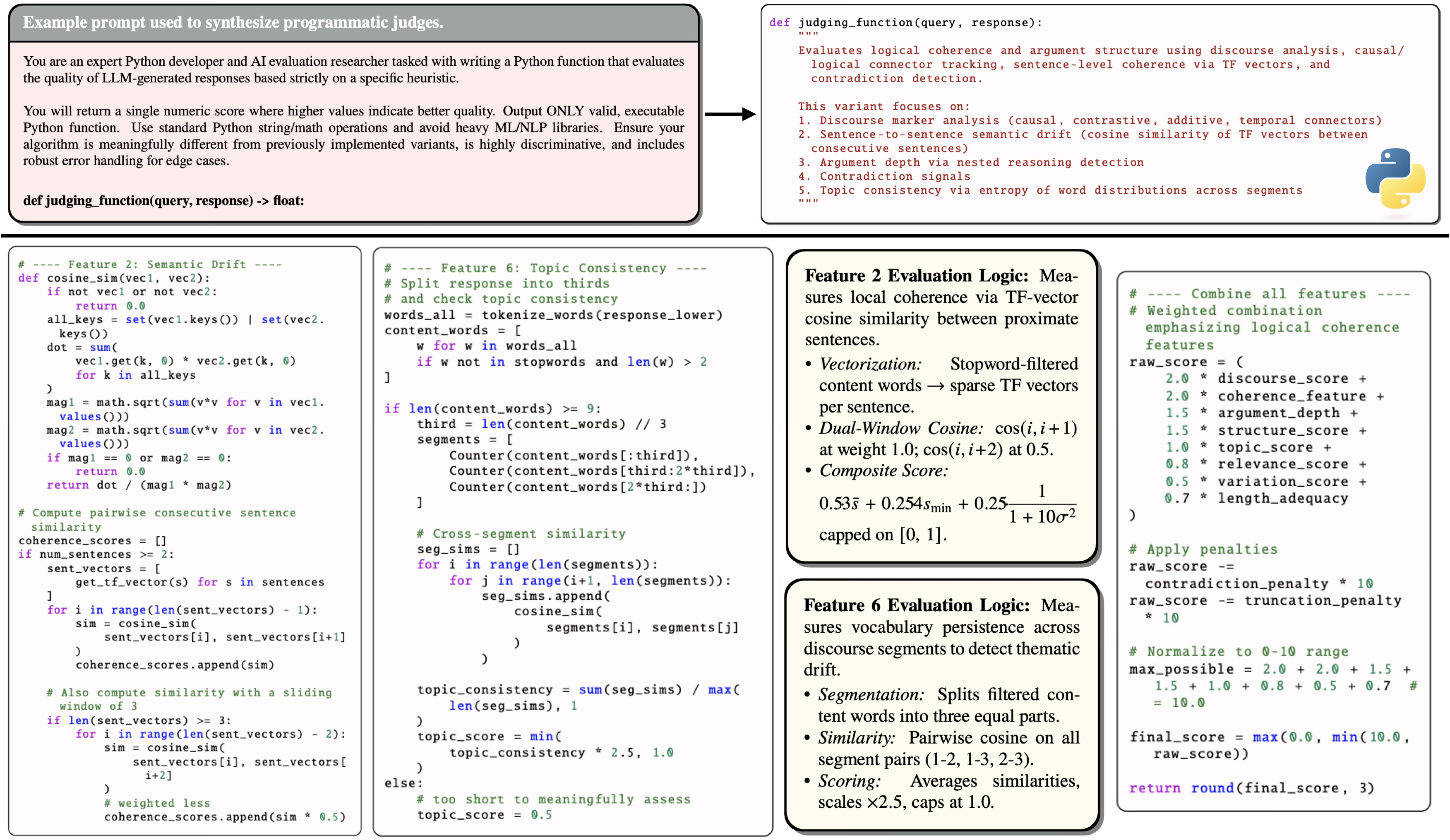}
    \caption{
    \textbf{An example of synthesizing programmatic judges}.
    Given a synthesis prompt (top-left), an LLM generates a Python \texttt{judging\_function} that articulates an evaluation rubric---for example, logical coherence---using interpretable features such as semantic drift via TF-vector cosine similarity or topic consistency across discourse segments. 
    These features are combined into a weighted score, with explicit penalties for contradiction and truncation, where the weights are determined by the LLM itself.
    This yields a transparent, program-based evaluator distilled directly from an LLM's logic.
    }
    \label{fig:demonstration}
\end{figure*}

This strategy offers three immediate advantages.
\textbf{First}, API costs now scale with the number of generated programs (tiny) rather than the size of the dataset (often huge): once synthesized, programs can be stored and reused locally at no additional API cost.
\textbf{Second}, program execution can be orders of magnitude faster than model inference, delivering extremely low-latency, high-throughput decision.
\textbf{Third}, programs are interpretable: practitioners or coding agents (e.g., Codex and Claude Code) can inspect each line, refine the judging logic, or incorporate domain knowledge---turning a model-based assessment into a transparent, and potentially formally verifiable procedure.

However, programmatic judging introduces its own challenges. 
First, a single program rarely generalizes to every input.
Second, even with multiple programs, naïve synthesis tends to yield repetitive logic.
Third, programs generated from different rubrics produce noisy scores at different scales.
We address these with \textbf{\textsc{Pajama}} (\underline{P}rogram-\underline{A}s-a-\underline{J}udge \underline{A}utomated \underline{M}odel \underline{A}ssessment), a system that is built on three components: 
\textbf{(i)} a curated set of evaluation rubrics---each expressible as code---to steer synthesis using diverse decision criteria;
\textbf{(ii)} a modeling step that calibrates program outputs and resolves their conflicts into a joint verdict, and
\textbf{(iii)} a confidence-aware fallback that routes uncertain samples to an LLM judge, yielding a hybrid evaluation that combines the efficiency of programmatic judges with the reliability of LLM evaluation.

We validate \textsc{Pajama} on five preference datasets spanning eight model families.
We show that standalone programmatic judges can match mid-sized LLMs, for instance, they reach the accuracy of \textsc{OLMo-2-13B-Instruct}~\citep{walsh2025} while running $47.25\times$ faster.
When combined with an LLM through our confidence-aware router, \textsc{Pajama} further advances the accuracy--throughput Pareto frontier.
For example, when paired with \textsc{OLMo-2-7B-Instruct}, it improves $+5.0\%$ accuracy at $2.9\times$ throughput, $+2.8\%$ over \textsc{Gemma-4-E2B-It}~\citep{team2026gemma} at $2.6\times$ throughput, and $+2.6\%$ over \textsc{Qwen2.5-3B-Instruct}~\citep{Yang2024Qwen25TR} at $2.2\times$ throughput.
On \textsc{RewardBench}~\citep{lambert-etal-2025-rewardbench}, a reward model distilled from programmatic judges' labels outperforms one trained on a proprietary LLM's preferences at $50\times$ lower API cost---with zero proprietary calls at evaluation time.
Finally, we show that program-based evaluation is robust to biased samples, and that pairing it with a coding agent for iterative program calibration improves this robustness further.

We summarize our contributions as follows:
\begin{itemize}[nosep,topsep=0pt,leftmargin=*]
    \item \textbf{A New Evaluation Paradigm.} 
    We introduce an approach that distills LLM logic into a program committee, replacing per-sample inference with one-time synthesis followed by local execution.
    \item \textbf{The \textsc{Pajama} System.} 
    We develop \textsc{Pajama}, a hybrid system that synthesizes diverse programmatic judges, calibrates and aggregates their verdicts, then routes uncertain cases to an LLM fallback.
    \item \textbf{Routing on Program Verdicts.} 
    We show that program outputs themselves provide an efficient confidence signal, enabling a lightweight router that combines two evaluation systems effectively.
    \item \textbf{Advancing the Pareto Frontier.}
    Across eight model families, \textsc{Pajama} matches strong LLM judges at a fraction of the cost and pushes the accuracy--throughput Pareto frontier.
    \item \textbf{Cheap, High-Quality Reward Signals.} 
    Reward models distilled from programmatic judges outperform those trained on proprietary LLM labels on \textsc{RewardBench}, at $50\times$ lower cost.
\end{itemize}
\section{Related Work}
\label{sec:related_works}

\textsc{Pajama} intersects three lines of work: automated evaluation, weak supervision, and routing strategies.

\noindent \textbf{Automated Evaluation.}
One of the key breakthroughs of LLMs is their ability to replace or augment human annotators in providing automated evaluations~\citep{gu2024survey, li2024llms}.
Prior work has demonstrated that LLM judges produce reliable decisions that align with human preferences across tasks such as ranking, pairwise comparison, and rubric-based scoring~\citep{wang2024pandalm, chiang-lee-2023-large, dubois2024lengthcontrolled, miranda2025hybrid}.
More recent efforts integrate LLM judges into the post-training pipeline, where their evaluations are used to train reward models~\citep{ouyang2022training, Bai2022TrainingAH}, supply feedback for reinforcement learning~\citep{gunjal2026rubrics, huang2026rubicap}, or fine-tune continually for specialized judge models~\citep{zhu2025judgelm, kim2024prometheus}.
While effective, LLM-based evaluation incurs substantial inference costs and inherits biases from pretraining data and prompt design, raising concerns about scalability and reliability~\citep{chen-etal-2024-humans, ye2025justice, shi-etal-2025-judging, wang-etal-2025-improving-llm-judge, zhao2026care}.
Our work explores a new direction: \textbf{\emph{synthesizing programmatic judges that distill LLM's logic into low-cost, transparent, and flexible evaluators.}}

\noindent \textbf{Weak Supervision.}
Weak supervision enables the rapid creation of labeled datasets by aggregating multiple noisy label estimates~\citep{ratner2017snorkel, ratner2016data, ratner2019training, shin2022universalizing, ratner2018snorkel} from sources such as heuristic rules, domain knowledge, or pretrained models.
These estimates are typically encoded as labeling functions, whose pseudo-labels are modeled and combined into a single probabilistic labeling decision.
The weak supervision paradigm has demonstrated success across diverse domains~\citep{roberts2022autows, huang2024multimodal, hooper2021cut, arora2023ask, saad2025shrinking, huang2025evaluating, vishwakarma2022lifting}.
Most prior work focuses on label aggregation to construct classification datasets.
Our system, \textsc{PAJAMA}, adapts it for a new purpose: \textbf{\emph{modeling the verdicts of programmatic judges to reach a combined and a more reliable evaluation decision.}}

\noindent \textbf{Routing Strategies.}
LLM routing systems leverage the complementary strengths of diverse LLMs rather than committing to a single one~\citep{ong2025routellm, hu2024routerbench, ding2024hybrid}.
A router directs each query to an appropriate model based on factors such as task difficulty, expected accuracy, and inference budget~\citep{jitkrittum2026universal}.
Existing routing strategies can be broadly categorized into \emph{model-free} approaches---which rely on heuristics such as nearest-neighbor retrieval over similar examples~\citep{hu2024routerbench}---and \emph{model-based} approaches, which, for example, train a classifier (e.g., a fine-tuned BERT) to predict the best model for each query~\citep{shnitzer2023large, hari2023tryage, lu2024routing}.
The former is limited by \emph{the strength of its heuristics}, while the latter \emph{incurs additional inference latency and requires supervision for training.}
In this work, we propose a new routing strategy: we derive confidence directly from the outputs of programmatic judges and use this signal to selectively escalate uncertain examples to an LLM judge.
\textbf{\emph{This yields a model-free fallback mechanism that preserves the high throughput of programmatic evaluation while retaining the stronger reasoning capabilities of LLMs on difficult examples.}}
\section{Framework}
\label{sec:framework}

We begin with an overview of \textsc{Pajama}'s workflow, followed by the problem setup in \S\ref{subsec:setup}.
\S\ref{subsec:synthesis} describes how we distill LLM evaluation into programmatic judges, and \S\ref{subsec:processing} discusses how we model their program outputs into a final verdict.
Finally, \S\ref{subsec:routing} presents a fallback mechanism to handle cases that the programmatic judges cannot cover or are uncertain about.

\begin{figure*}[t!]
    \centering\includegraphics[width=\linewidth]{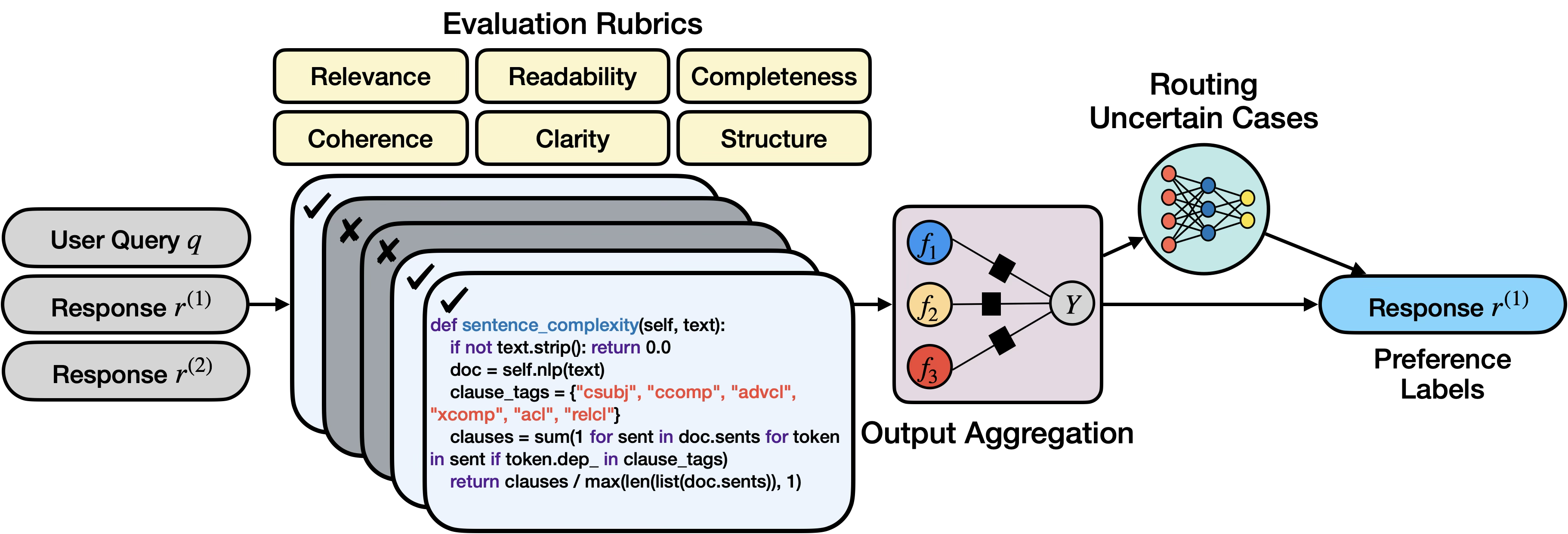}
    \caption{
    \textbf{\textsc{Pajama} Workflow.}
    Given a query $q$ and two candidate responses $r^{(1)}, r^{(2)}$, a diverse pool of programmatic judges---synthesized by an LLM from curated rubrics---produces initial evaluations.
    These program outputs are calibrated and selected, and their verdicts are aggregated into a combined decision.
    Uncertain cases are then routed to an LLM judge to produce the final preference.
    }
    \label{fig:framwork_figure}
\end{figure*}

\noindent \textbf{General Workflow.}
Figure~\ref{fig:framwork_figure} illustrates the \textsc{Pajama} workflow.
Given a dataset of queries paired with two candidate responses, we first prompt an LLM to synthesize Python programs that encode different judging rubrics; varying the prompt and evaluation criteria yields a diverse pool of programmatic judges.
We then calibrate each program's outputs using a held-out validation set, select the most effective programs, and aggregate their verdicts into a single preference decision.
For inputs that are uncovered or yield low confidence, an efficient routing mechanism falls back to an LLM judge.

\subsection{Problem Setup}
\label{subsec:setup}

Let $\mathcal{Q}, \mathcal{R} \subseteq \Sigma^\star$ denote the spaces of queries and responses, where $\Sigma^\star$ is the space of free-form text.
Given a query $q \in \mathcal{Q}$ and two candidate responses $r^{(1)}, r^{(2)} \in \mathcal{R}$, produced by the same or different LLMs, our goal is to determine which response is preferred under a target notion of quality (e.g., alignment with human preferences).
\subsection{Distilling into Programmatic Judges}
\label{subsec:synthesis}

Model-based evaluators, such as LLM-as-a-judge, incur high inference cost and are opaque in their decision-making process.
To address these, we translate LLM judging logic into programs that assess $q$, $r^{(1)}$, and $r^{(2)}$ directly.
We design a prompt template that instructs an LLM to synthesize Python functions, which serve as our judges.
Each function is asked to take a query and a response as direct inputs, returns a scalar score, where a higher value indicates higher quality according to its articulated evaluation logic.
Figure~\ref{fig:demonstration} shows a simplified example of our prompt and the program it generates.

A naïve synthesis approach often yields repetitive programs.
To encourage diversity, we curate ten distinct evaluation rubrics, each of which can be expressed as an executable program.
For each synthesis round, we select a rubric and incorporate it into the prompt, together with few-shot exemplars from the target dataset, to guide program generation.
We then perform a text-based similarity check to filter out newly generated programs whose evaluation logic is similar to that of an existing program in the pool.
We adopt this approach for its simplicity, but note that \emph{more sophisticated program synthesis techniques or customizations can easily be swapped in}.
Appendix~\ref{app:prompt_collection} provides the complete synthesis prompt, our curated rubrics, and details on their construction.
\subsection{Modeling Programmatic Judges}
\label{subsec:processing}

Next, we present our modeling procedure for converting program outputs into a reliable verdict.

\noindent \textbf{Program Output Calibration.}
Each synthesized program acts as an independent judge $f_j$ that scores the quality of LLM responses.
Given a tuple $(q_i, r_i^{(1)}, r_i^{(2)})$, we execute each program to obtain quality scores $s_{ij}^{(m)} := f_j(q_i, r_i^{(m)})$ for $m \in \{1,2\}$.
We then apply min-max normalization to map each program's outputs into $[0,1]$ (since programs may use different scales).
Using these normalized scores $\hat{s}_{ij}^{(1)}$ and $\hat{s}_{ij}^{(2)}$, we compute the \emph{quality difference} $d_{ij} := \hat{s}_{ij}^{(1)} - \hat{s}_{ij}^{(2)} \in [-1, 1]$, where positive values favor $r_i^{(1)}$.
We convert this difference into a discrete verdict $v_{ij} \in \{-1, 0, +1\}$ using a per-program threshold $\tau_j \geq 0$: $v_{ij} = +1$ if $d_{ij} > \tau_j$, $v_{ij} = -1$ if $d_{ij} < -\tau_j$, and $v_{ij} = 0$ (abstain) otherwise.
This abstention margin allows each program to \emph{withhold its vote when its own signal is too weak to be trusted}.
When a validation set is available (possibly small), we set each $\tau_j$ to the value that maximizes program $j$'s accuracy on that set and reuse these thresholds at inference time\footnote{A validation set is not strictly required: when none is available, we simply set $\tau_j = 0$, which reduces the verdict to a direct comparison between $\hat{s}_{ij}^{(1)}$ and $\hat{s}_{ij}^{(2)}$.}.

\noindent \textbf{Top-$k$ Program Selection.}
Not all synthesized programs are equally reliable.
When a validation set is available, we discard programs with below random chance accuracy ($<50\%$) and keep the top-$k$ remaining programs by validation accuracy as the final committee.

\noindent \textbf{Program Verdict Aggregation.}
Individual program verdicts can be noisy and may disagree.
At the same time, programs based on different rubrics can also capture complementary aspects of response quality.
We therefore aggregate them to reduce noise and obtain a more reliable preference.
Specifically, we apply aggregation step in weak supervision literature~\citep{ratner2017snorkel, ratner2016data, ratner2019training, shin2022universalizing, ratner2018snorkel}, which models noisy votes into higher-quality \emph{pseudolabel} by using a generative model to estimate each program's accuracy from their agreement and disagreement patterns.
We collect the top-$k$ programs' votes on $n$ validation samples into a preference matrix $\mathbf{L} \in \{-1, 0, +1\}^{n \times k}$, then apply a standard label model (e.g., from the Snorkel framework~\citep{ratner2017snorkel, ratner2016data, ratner2018snorkel}) to learn program-specific aggregation weights.
At inference time, these learned weights are applied to obtain an aggregated preference.
\subsection{Routing Uncertain Cases}
\label{subsec:routing}

Although aggregating programmatic judges improves reliability, some inputs remain difficult to evaluate confidently.
For example, all selected programs may abstain ($v_{ij}=0$ for all $j$), or the aggregator may assign similar probability to both candidate responses, resulting in a posterior close to $0.5$.
Rather than forcing the program committee to make an uncertain decision, we use signals derived directly from program outputs---such as vote disagreement or aggregated confidence---to identify these cases and route them to an LLM judge.
This yields a hybrid evaluation system that combines the best of both worlds: \textit{\textbf{programmatic judges deliver quick verdicts, while uncovered or low-confidence cases are routed to an LLM judge, trading a small fraction of expensive calls for improved reliability (i.e., evaluation accuracy).}}

\section{Experiments}
\label{sec:experiments}

We evaluate \textsc{Pajama} across four different settings, each designed to validate programmatic judges' benefits.
Through them, we confirm four key claims:
\begin{enumerate}[nosep,topsep=0pt,leftmargin=*,label=\textbf{C\arabic*.}, leftmargin=*]
    \item 
    \textbf{Accuracy and Throughput.} 
    Programmatic judges can match the accuracy of mid-sized LLM judges while achieving substantially higher evaluation throughput than standard LLM inference.
    \item 
    \textbf{Routing and Pareto Frontier.}
    Program-derived signals enable an effective router; when combined with an LLM, \textsc{Pajama} significantly advances the accuracy--throughput Pareto frontier.
    \item 
    \textbf{Cost-effective Distillation.}
    Programmatic judges provide cost-effective supervision for reward model distillation, achieving higher performance than labels from proprietary LLMs.
    \item
    \textbf{Robustness to Bias.}
    Programs exhibit resilience to systemic biases comparable to a 7B LLM; moreover, automated program calibration via coding agents further enhances this robustness.
\end{enumerate}

\subsection{Effectiveness of Program-based Evaluation}
\label{subsec:efficient_evaluator}

\begin{figure*}[t!]
    \centering\includegraphics[width=\linewidth]{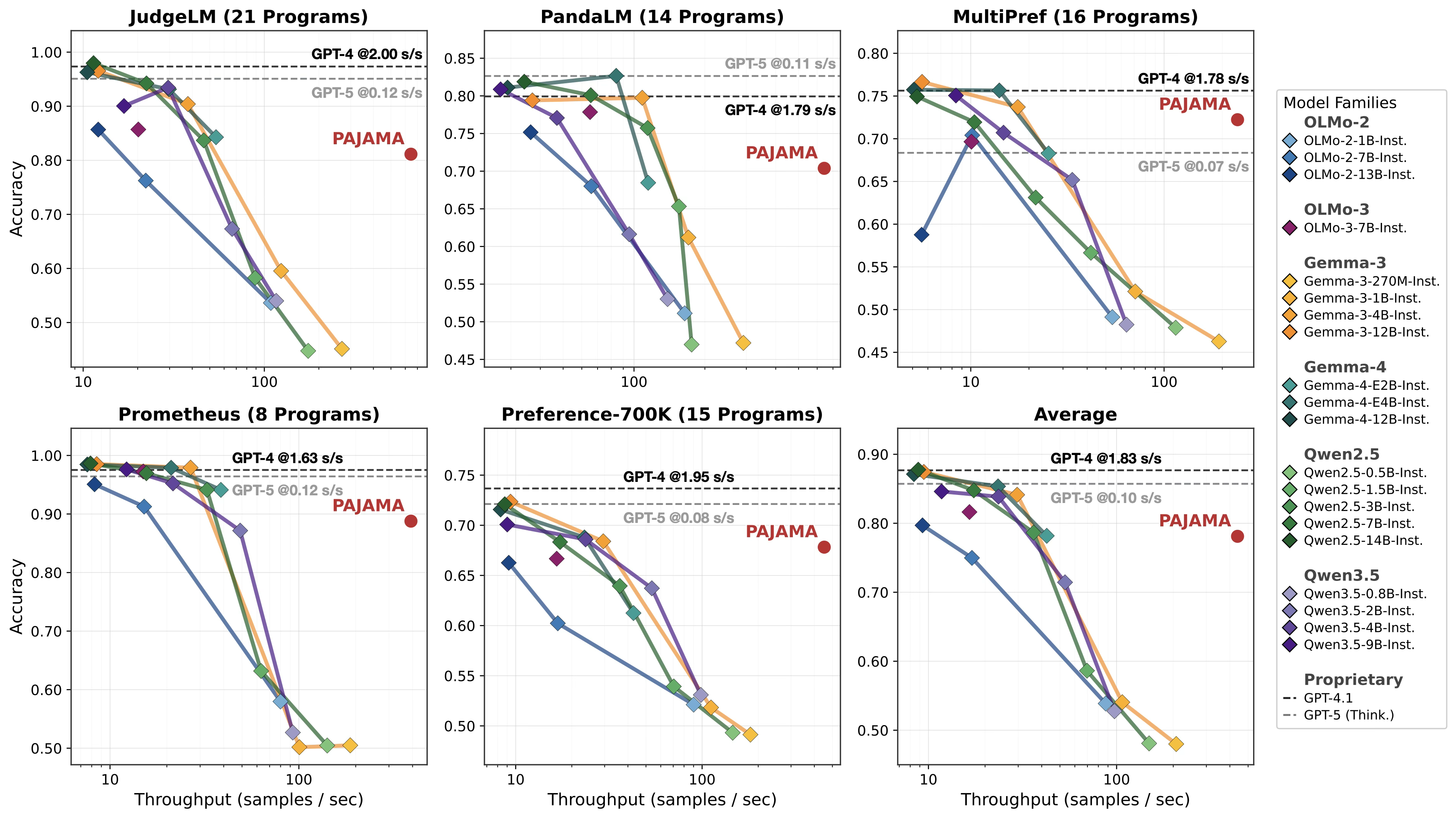}
    \caption{\textbf{Effectiveness of Programmatic Judges.}
    Accuracy vs.\ throughput across five preference datasets.
    Each subplot title reports the number of programs retained after selection.
    Programmatic judges achieve \textbf{\emph{extremely high throughput}} while remaining competitive in accuracy with mid-sized LLM judges, e.g., matching \textsc{OLMo-2-13B-Instruct} on average.
    }
    \label{fig:pajama_effectiveness}
\end{figure*}

Programmatic judges distill LLM evaluation logic into executable programs that produce pairwise preference judgments.
We first ask whether this design is both \emph{accurate}---in terms of evaluation performance---and \emph{fast}, in terms of throughput.

\noindent \textbf{Setup.}
We evaluate on five pairwise preference datasets: JudgeLM~\citep{zhu2025judgelm}, PandaLM~\citep{wang2024pandalm}, MultiPref~\citep{miranda2025hybrid}, Prometheus~\citep{kim2024prometheus}, and Preference-700K~\citep{dong2024rlhf}.
From each, we sample up to 5{,}000 examples for evaluation and hold out an additional 500 examples as a validation set for modeling program outputs (i.e., calibration, selection, and aggregation).
We make a one-time investment to synthesize 80 candidate programs with Claude Opus 4.6~\citep{anthropic2026claudeopus46systemcard}, using in-context prompts seeded with 10 examples randomly drawn the validation set.
On samples where the program committee abstains, we assign labels randomly so that coverage is complete\footnote{As shown in Table~\ref{tab:main}, most of selected programs yield high coverage (e.g., $>95.0\%$); that is, only a few uncovered samples are annotated by random guessing.}.
Appendices~\ref{app:prompt_collection} and~\ref{app:dataset_description} provide the full synthesis prompts, curated rubrics, and dataset descriptions.
Appendix~\ref{app:ablation_rubrics} ablates each design choice in \textsc{Pajama}, including \textbf{(i)} rubric contribution, \textbf{(ii)} the program synthesizer, \textbf{(iii)} program committee size, \textbf{(iv)} validation set size, and \textbf{(v)} the cross-dataset generalization.

We compare this program-based evaluation against eight model families spanning a wide range of scales: proprietary judges (GPT-4.1~\citep{achiam2023gpt}, GPT-5 with Thinking~\citep{singh2025openai}), \textsc{OLMo-2}~\citep{walsh2025}, \textsc{OLMo-3}~\citep{olmo2025olmo}, \textsc{Gemma-3}~\citep{kamath2025gemma}, \textsc{Gemma-4}~\citep{team2026gemma}, \textsc{Qwen2.5}~\citep{Yang2024Qwen25TR}, and \textsc{Qwen3.5}~\citep{team2026qwen3}.
We measure two quantities: \emph{accuracy} against ground-truth preference labels, and \emph{throughput}, the number of samples judged per second.
LLM judges are served with vLLM~\citep{kwon2023efficient}, a widely adopted inference engine, using 64 concurrent requests.
Programmatic judges are executed in parallel across 24 CPU threads, and the reported throughput accounts for the time needed to aggregate the program outputs.

\noindent \textbf{Results.}
Figure~\ref{fig:pajama_effectiveness} presents accuracy against throughput on each dataset and on their average; full numerical results are reported in Appendix~\ref{subsec:full_table}.
Two findings stand out.
\textit{\textbf{First, programmatic judges achieve competitive accuracy with mid-sized LLM judges.}}
On average, they attain an accuracy of $78.11\%$, on par with \textsc{OLMo-2-13B-Instruct}, \textsc{Gemma-4-E2B-It}, and \textsc{Qwen2.5-3B-Instruct}, and within 8 points of the strongest proprietary judge, GPT-5 Thinking ($85.72\%$).
On Prometheus, a committee of only eight programs reaches $88.78\%$, matching \textsc{OLMo-2-7B-Instruct}.
\textit{\textbf{Second, program-based evaluation occupies a throughput regime that no LLM judge can reach.}}
It runs $2.12\times$ faster than \textsc{Gemma-3-270M-It}---the smallest model we test---while outperforming it by $+30.11$ accuracy points.
Against larger variants (e.g., \textsc{Qwen2.5-14B-Instruct}, \textsc{Gemma-4-12B-It}), programs run roughly $50\times$ faster while approaching their accuracy.
These results demonstrate that \textbf{\textit{a committee of fewer than twenty synthesized programs is sufficient to deliver competitive accuracy while running orders of magnitude faster than any LLM judge we tested.}}

\subsection{Hybrid Evaluation Advances The Pareto Frontier}
\label{subsec:pareto_frontier}

\begin{figure*}[t!]
    \includegraphics[width=\linewidth]{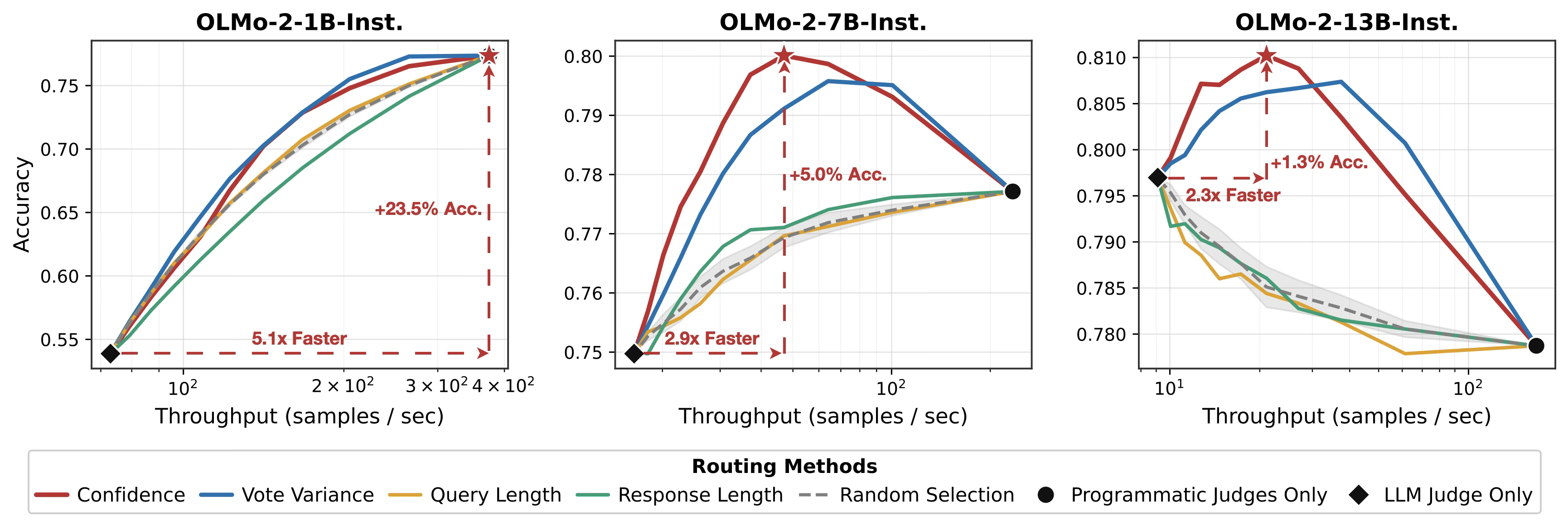}
    \centering
    \caption{
    \textbf{Routing within the OLMo-2 family.} 
    Accuracy vs.\ throughput as the escalation threshold is swept across different routing signals.
    Endpoints represent pure-program (right-most) and pure-LLM (left-most) evaluation.
    \textsc{Pajama}-derived signals (\emph{aggregator posterior}, \emph{vote variance}) consistently dominate other model-free routing methods across all model sizes.
    }
    \label{fig:olmo2_result}
\end{figure*}

In \S\ref{subsec:efficient_evaluator}, we show that standalone programmatic judges dominate a significant region of the accuracy--throughput Pareto frontier.
We next ask whether programmatic and LLM judges can be combined to exploit their complementary strengths.
\emph{What if} programs handled clear-cut pairs reliably, routing only the uncertain cases to an LLM?
This should recover accuracy while preserving high throughput.

\noindent \textbf{Setup.}
We validate this idea with a \textbf{\textit{staged evaluation policy: programs assess every pair, and only uncovered or low-confidence ones are escalated}}.
We identify uncertain cases using two signals derived from the program committee:
\textit{\textbf{(i) vote variance}}---the disagreement among synthesized programs, where \emph{high} variance indicates uncertainty; and
\textit{\textbf{(ii) aggregator posterior}}---the posterior probability inferred by the aggregator, where a probability close to $0.5$ indicates uncertainty.
We use a threshold to control the escalation rate: samples flagged as low-confidence are routed, while the rest are decided by the programmatic judges.
Sweeping the threshold traces the full hybrid accuracy--throughput curve, with the two endpoints recovering programs-only and LLM-only evaluation.

We compare this strategy against three heuristic routing baselines that do not leverage \textsc{Pajama}'s signals:
\textit{\textbf{query length}}, which escalates pairs with longer prompts;
\textit{\textbf{response length}}, which escalates pairs with longer candidate responses;
and \textit{\textbf{random selection}}, which routes a uniformly sampled fraction of pairs.
Each baseline is evaluated across the same range of escalation rates.
We report average testing accuracy across the five preference datasets.

\noindent \textbf{Results.}
We present the analysis in two parts.
First, we compare routing signals within a single model family to identify which signal works best.
Then, we apply the best signal across nineteen LLM judges to characterize how the hybrid frontier moves relative to pure-LLM evaluation.

\textit{\textbf{First, program-derived routing signals improve evaluation quality.}}
Figure~\ref{fig:olmo2_result} compares routing strategies on the \textsc{OLMo-2} family.
Across all three sizes, both the aggregator posterior (the prediction confidence) and vote variance trace curves that dominate the length-based and random baselines: \textit{\textbf{at any throughput they yield higher accuracy, and at any accuracy they yield higher throughput.}}
For example, with \textsc{OLMo-2-7B-Instruct}, the hybrid assessment routed by aggregator posterior achieves a $+5.0\%$ accuracy improvement at $2.9\times$ higher throughput than LLM-only evaluation.
Appendix~\ref{subsec:fallback_rate} further reports a fallback-rate sensitivity analysis.
For most hybrid configurations with small LLMs, accuracy peaks at fallback rates between $32\%$ and $42\%$. 
This means \textsc{Pajama} can exceed the accuracy of LLM-only evaluation while escalating fewer than half of the samples to the LLM.
Additional routing sweeps across other model families are provided in Appendix~\ref{subsec:routing_with_judges}.
Finally, \textit{\textbf{\textsc{Pajama}'s fallback mechanism relies entirely on internal signals}}, making its routing decisions incur negligible overhead---unlike model-based routers, which typically require an extra model inference~\citep{hu2024routerbench, shnitzer2023large, hari2023tryage, lu2024routing}.
We present this comparison in Appendix~\ref{subsec:model_based_router}, where program-based routing matches an oracle-supervised model-based router with no supervision or additional latency.

\begin{figure*}[t!]
    \centering\includegraphics[width=\linewidth]{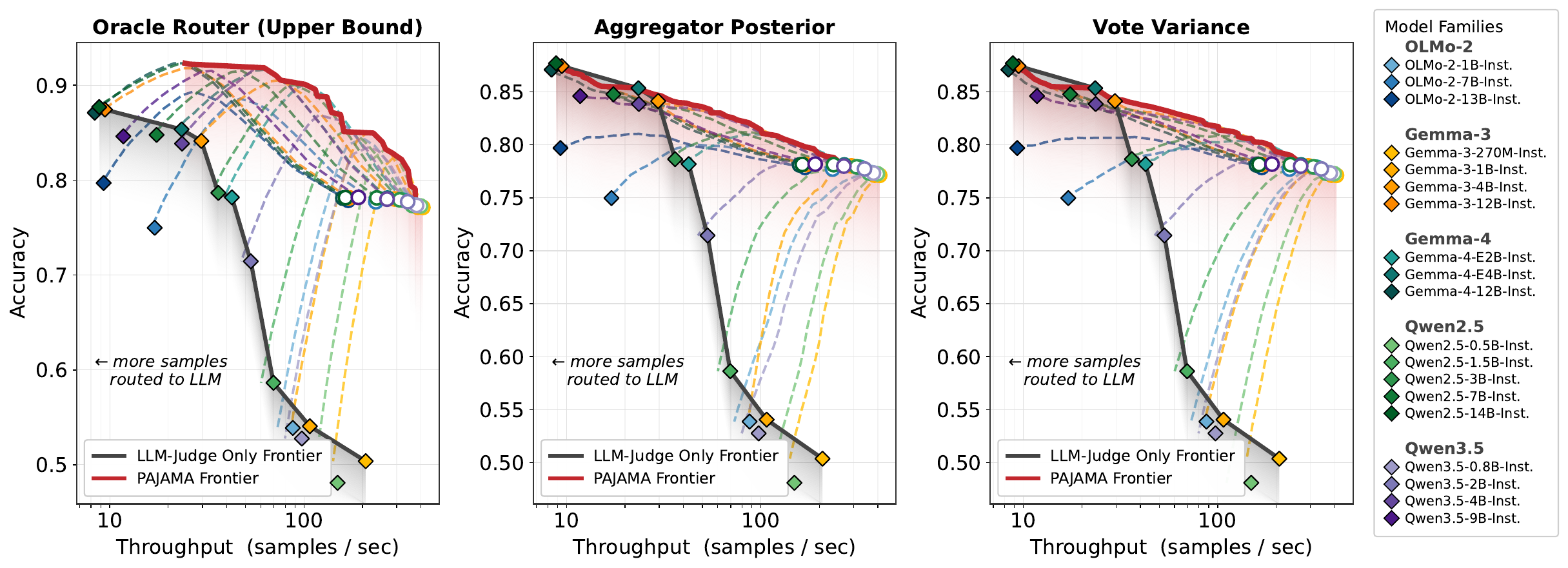}
    \centering\caption{
    \textbf{Hybrid evaluation pushes the Pareto frontier further.}
    Each panel shows accuracy vs.\ throughput as \textsc{Pajama} routes uncertain pairs to one of nineteen LLM judges.
    Dashed curves trace each judge's hybrid trajectory;
    the {\color{red!75} \textbf{red envelope}} is the resulting \textsc{Pajama} frontier, while the black envelope is the LLM-only one from \S\ref{subsec:efficient_evaluator}.
    }
    \label{fig:pareto_curve}
\end{figure*}

\emph{\textbf{Second, hybrid evaluation extends the Pareto frontier.}}
Figure~\ref{fig:pareto_curve} demonstrates this.
Each dashed line traces one judge's hybrid trajectory as the threshold is swept; the red envelope is the resulting frontier with \textsc{Pajama}'s routing, and the black envelope is the LLM-only frontier from \S\ref{subsec:efficient_evaluator}.
Across all nineteen judges, the red envelope dominates the black one: \textit{\textbf{at every throughput level, hybrid evaluation achieves higher accuracy}}.
For example, routing uncertain cases to \textsc{OLMo-3-7B-Instruct} yields $82.20\%$ accuracy at $2.19\times$ the throughput of the LLM-only system, while \textsc{GPT-5 (Thinking)} reaches $85.72\%$ accuracy at only $0.10$ samples/sec.
\emph{\textbf{Our hybrid configurations provide competitive reliability while running hundreds of times faster and incurring zero proprietary API cost.}}

In the leftmost panel, we further present an \emph{oracle router} that escalates only the pairs the programmatic judges would have mislabeled, representing the upper bound of any routing strategy built on our programs.
The remaining gap between our internal signals and the oracle suggests that richer signals from the program committee---beyond vote variance and posterior probability---could further improve routing.
\textbf{\emph{Overall, this Pareto frontier shift allows practitioners to obtain higher accuracy under a fixed throughput budget, or substantially higher throughput at a target accuracy, without training stronger models or increasing API spend.}}
\subsection{Reward Model Distillation at Lower Cost}
\label{subsec:reward_model}

LLM judges are widely used not only for assessment but also for generating preference labels that enable reward model distillation with broader generalization.
We ask whether program-based verdicts can serve as an effective training signal in the post-training stage.

\noindent \textbf{Setup.}
We sample an additional 20{,}000 preference pairs from Prometheus and JudgeLM to build training sets, relabel them with programmatic judges, and compare against the original GPT-4 labels~\citep{achiam2023gpt} provided by these two datasets.
We then fine-tune \textsc{Qwen2.5-3B-Instruct} on each supervision source under the standard Bradley--Terry objective~\citep{dong2024rlhf}.
We evaluate along two axes: (i) in-domain accuracy on the Prometheus and JudgeLM test sets, and (ii) out-of-domain generalization on \textsc{RewardBench}~\citep{lambert-etal-2025-rewardbench}.
We also report the estimated API cost of each labeling source: GPT-4 inference over all 20{,}000 pairs versus our one-time cost of synthesizing the 80 programs.
Our training configurations and per-category \textsc{RewardBench} results are presented in Appendices~\ref{app:experimental_details} and~\ref{subsec:reward_model_performance}.

\noindent \textbf{Results.}
Table~\ref{tab:reward_model_distillation} reports reward model performance and labeling costs.
\textit{\textbf{Programmatic judges provide a more cost-effective training signal than proprietary GPT-4 labels.}}
In-domain, our labels yield competitive accuracy at roughly $50\times$ lower API cost, reaching $92.20\%$ on Prometheus.
Out-of-domain, reward models trained on \textsc{Pajama} labels transfer better to \textsc{RewardBench}, improving the average score by $+3.60$ points on Prometheus and $+4.05$ points on JudgeLM, with the largest gains on the Reasoning and Safety categories.
\textit{\textbf{Moreover, since our program synthesis is a one-time investment ($\mathcal{O}(1)$), the spending gap widens as the dataset grows, whereas GPT-4 labeling scales linearly ($\mathcal{O}(n)$) with the number of pairs}}.
This makes \textup{\textsc{Pajama}} an attractive labeling source for alignment workflows with limited distillation budgets.

\begin{table}[t]
\centering
\caption{
\textbf{Reward model distillation with \textsc{Pajama} vs.\ GPT-4 labels.}
\textsc{Pajama} matches or surpasses GPT-4 supervision on \textsc{RewardBench} on average while reducing labeling cost by $45$--$50\times$.
}
\vspace{-8pt}
\small
\setlength{\tabcolsep}{3pt}
\renewcommand{\arraystretch}{1.3}
\sisetup{detect-weight=true, detect-shape=true, detect-mode=true}
\resizebox{0.95\textwidth}{!}{
\begin{tabular}{lccccccccc}
\toprule
\multirow{2}{*}{\textbf{Labeling Source}} & \multirow{2}{*}{\textbf{API Calls}} & \multirow{2}{*}{\textbf{Cost Scaling}} & \multirow{2}{*}{\textbf{Estimated Cost}} & \textbf{In-domain} & \multicolumn{4}{c}{\textbf{RewardBench}} & \multirow{2}{*}{\textbf{Average}} \\
\cmidrule(lr){6-9}
 & & & & \textbf{Testing Acc.} & \textbf{Chat} & \textbf{Chat Hard} & \textbf{Reasoning} & \textbf{Safety} & \\
\midrule
\rowcolor{family}
\multicolumn{10}{l}{\textit{Trained on Prometheus Dataset}} \\
GPT-4 & 20{,}000 samples & $\mathcal{O}(n)$ & \$363.97 & \textbf{97.23} & 68.44 & \textbf{38.82} & 55.70 & 45.95 & 52.23 \\
\rowcolor{pajamablue}
\textbf{PAJAMA} & \textbf{80} programs & $\mathcal{O}(1)$ & \textbf{\$7.21} \textbf{\color{purple}{(50$\times$ cheaper)}} & 92.20 & \textbf{79.33} & 30.04 & \textbf{60.58} & \textbf{53.38} & \textbf{55.83} \\
\midrule
\rowcolor{family}
\multicolumn{10}{l}{\textit{Trained on JudgeLM Dataset}} \\
GPT-4 & 20{,}000 samples & $\mathcal{O}(n)$ & \$296.37 & \textbf{90.24} & \textbf{67.88} & 38.82 & 65.13 & 58.24 & 57.52 \\
\rowcolor{pajamablue}
\textbf{PAJAMA} & \textbf{80} programs & $\mathcal{O}(1)$ & \textbf{\$6.49} \textbf{\color{purple}{(45$\times$ cheaper)}} & 82.79 & \textbf{67.88} & \textbf{44.52} & \textbf{72.91} & \textbf{60.95} & \textbf{61.57} \\
\bottomrule
\end{tabular}
}
\label{tab:reward_model_distillation}
\end{table}

\subsection{Bias Reduction and Program Calibration}
\label{subsec:bias_reduction}

LLM judges sometimes rely on superficial features rather than the true quality when making decisions.
Programmatic judges offer a transparent alternative.
We ask whether this transparency yields \emph{improved robustness}, and whether programs can be \emph{easily calibrated} when biases are detected.

\noindent \textbf{Setup.}
We study five well-known bias types:
\textit{\textbf{(i) position bias}}, favoring answers based on their order;
\textit{\textbf{(ii) rich-content}}, favoring formatting (e.g., markdown, emojis) over factual accuracy;
\textit{\textbf{(iii) reference bias}}, crediting unsupported citations;
\textit{\textbf{(iv) gender bias}}, reacting to gendered phrasing unrelated to quality;
and \textit{\textbf{(v) verbosity bias}}, favoring longer responses regardless of substantive value.

In each case, the two candidate responses are matched in quality, so there is no inherent winner.
We compare each judge's decision before and after a controlled perturbation of one response.
For \textbf{\textit{position bias}}, we query the judge twice with the candidate order swapped.
For the other four \textbf{\textit{content biases}}, we run a clean trial and a perturbed trial, measuring whether the perturbation flips the preference toward the biased response.
For \textsc{Pajama}, we aggregate the verdicts of 8 selected programs from Prometheus.
We assess robustness with two metrics: \textbf{\textit{Flip Rate (FR)}}, the percentage of samples whose verdict changes under perturbation, and \textbf{\textit{Bias Win Rate (BWR)}}, the percentage in which the biased response ultimately wins.
Both metrics are lower-is-better.
Appendix~\ref{app:dataset_description} details the data sources.

\begin{table}[t!]
\centering
\caption{
\textbf{Bias Robustness.}
Flip Rate (FR) and Bias Win Rate (BWR); \emph{lower is better}.
LLM results are averaged over three trials.
\textsc{Pajama} has the lowest average FR, and its calibrated variant, which applies targeted edits to the synthesized programs, lowers both metrics further.
}
\vspace{-8pt}
\label{tab:bias_reduction}
\small
\setlength{\tabcolsep}{3pt}
\renewcommand{\arraystretch}{1.3}
\sisetup{detect-weight=true, detect-shape=true, detect-mode=true}
\resizebox{0.9\textwidth}{!}{
\begin{tabular}{l c cc cc cc cc | cc}
\toprule
\multirow{2}{*}{\textbf{Method}} & \textbf{Position} & \multicolumn{2}{c}{\textbf{Rich Content}} & \multicolumn{2}{c}{\textbf{Reference}} & \multicolumn{2}{c}{\textbf{Gender}} & \multicolumn{2}{c}{\textbf{Verbosity}} & \multicolumn{2}{c}{\textbf{Average}} \\
\cmidrule(lr){2-2} \cmidrule(lr){3-4} \cmidrule(lr){5-6} \cmidrule(lr){7-8} \cmidrule(lr){9-10} \cmidrule(lr){11-12}
& \textbf{FR (\%)} & \textbf{FR (\%)} & \textbf{BWR (\%)} & \textbf{FR (\%)} & \textbf{BWR (\%)} & \textbf{FR (\%)} & \textbf{BWR (\%)} & \textbf{FR (\%) } & \textbf{BWR (\%)} & \textbf{FR (\%)} & \textbf{BWR (\%)} \\
\midrule
\rowcolor{family}
\multicolumn{12}{l}{\textit{OLMo-2-Instruct family}} \\
\quad 1B    & 59.36 & 17.94 & 61.73 & 21.87 & 71.22 & 18.93 & 65.25 & 21.16 & 32.69 & 27.85 & 57.72 \\
\quad 7B    & 82.82 & 10.99 & 14.08 & 26.62 & 31.55 & 4.51  & 7.32  & 18.47 & 57.53 & 28.68 & 27.62 \\
\quad 13B   & 93.52 & 1.83  & 89.58 & 2.11  & 77.32 & 1.55  & 50.99 & 1.67  & 2.00  & 20.14 & 54.97 \\
\midrule
\rowcolor{family}
\multicolumn{12}{l}{\textit{Gemma-3-Instruct family}} \\
\quad 4B   & 60.14 & 15.21 & 68.73 & 13.10 & 52.82 & 0.56  & 22.25 & 0.20  & 0.33  & 17.84 & 36.03 \\
\quad 12B  & 44.08 & 10.00 & 47.04 & 12.54 & 48.87 & 0.42  & 15.77 & 1.27  & 1.47  & 13.66 & 28.29 \\
\midrule
\rowcolor{family}
\multicolumn{12}{l}{\textit{Qwen2.5-Instruct family}} \\
\quad 1.5B & 66.48 & 15.77 & 58.17 & 17.32 & 62.96 & 4.65  & 49.44 & 41.13 & 61.47 & 29.07 & 58.01 \\
\quad 3B   & 55.77 & 17.89 & 65.07 & 11.83 & 61.13 & 0.42  & 21.83 & 12.27 & 29.20 & 19.63 & 44.31 \\
\quad 7B   & 46.48 & 19.86 & 65.77 & 13.10 & 45.63 & 0.99  & 17.61 & 11.20 & 19.60 & 18.33 & 37.15 \\
\quad 14B  & 40.00 & 16.06 & 42.39 & 18.17 & 42.82 & 0.99  & 5.49  & 2.87  & 7.87  & 15.61 & 24.64 \\
\midrule
\rowcolor{pajamablue} 
\textbf{PAJAMA}         & 0.00  & 23.13 & 48.92 & 20.30 & 47.10 & 7.69  & 46.27 & 9.34  & 16.84 & 12.09 & 39.78 \\
\rowcolor{pajamablue}
\textbf{PAJAMA (Calibrated)} & 0.00  & 13.74 & 53.28 & 17.29 & 43.17 & 3.70  & 36.17 & 3.25  & 9.68  & 7.60  & 35.58 \\
\rowcolor{red!15}
\textbf{$\Delta$ (Reduction)} & 0.00 & $-$9.39 & $+$4.36 & $-$3.01 & $-$3.93 & $-$3.99 & $-$10.10 & $-$6.09 & $-$7.16 & $-$4.49 & $-$4.20 \\
\bottomrule
\end{tabular}
}
\end{table}

\noindent \textbf{Results.}
Table~\ref{tab:bias_reduction} reports the robustness of LLM judges and programmatic judges across the five bias types.
On average, \textit{\textbf{\textup{\textsc{Pajama}} achieves a lower flip rate than all three families of LLM judges, while its bias win rate is comparable to \textup{\textsc{Qwen2.5-7B-Instruct}} and \textup{\textsc{Gemma-3-4B-Instruct}}}}.
For position bias, a program's reasoning is invariant to candidate order, yielding the highest consistency overall---a property that follows naturally from the design of the programs themselves.
For verbosity bias, the synthesized programs encode logic that penalizes responses that are either too short or too long, yielding low flip rates and bias win rates that outperform 7B models.

Program-based evaluation exposes the full decision process, and its judging logic can be edited through minor code changes.
We further use a coding agent (Claude Code) to calibrate the programmatic judges by making them aware of specific bias types; the calibration prompt is provided in Appendix~\ref{subsec:bias_prompt}.
Table~\ref{tab:bias_reduction} confirms the effectiveness of this calibration:
\textit{\textbf{both flip rate and bias win rate can be substantially reduced by editing the programs themselves}}.
For example, in the rich-content category, the programs are calibrated to remove bonuses for numbered lists, bullets, and colon headers, improving the flip rate by $9.39\%$.
Unlike opaque LLM judges, these gains stem directly from the programs' transparent design, \textbf{\textit{demonstrating that programmatic judges are not only auditable but also fixable.
Bias is no longer a fixed property of the evaluator, but a bug that can be patched}}.

\section{Conclusion}
\label{sec:conclusion}

We present \textsc{Pajama}, which distills LLM evaluation logic into programmatic judges---Python functions synthesized from curated rubrics, calibrated, and aggregated via weak supervision, with routing to an LLM judge when the committee is uncertain.
Across five preference datasets and eight model families, \textsc{Pajama} matches mid-sized LLM judges in accuracy at orders-of-magnitude higher throughput, advancing the Pareto frontier of LLM judges and producing labels that are more cost-effective than proprietary supervision for reward model distillation and robust to biased samples.

\bibliography{reference}
\bibliographystyle{achemso}

\newpage
\appendix
\section*{Appendix Roadmap}

Our appendix is organized as follows.
We begin with the prompts used in our framework and experiments: 
Appendix~\ref{app:prompt_collection} presents the complete instruction used to synthesize programmatic judges, along with the curated evaluation rubrics that guide generation.
Appendix~\ref{app:dataset_description} describes the datasets, including our data splits, their ground-truth sources, and the origin of the biased samples.
Appendix~\ref{app:experimental_details} details our experimental setup for reward model distillation, as well as the compute resources used.

We then turn to additional empirical results in Appendix~\ref{app:experimental_results}, where we report the full evaluation tables, \textbf{(ii)} fallback rate analysis, results for routing programmatic judges to an LLM judge from a different model family, comparison to a model-based router, and reward model performance on \textsc{RewardBench}.
Next, Appendix~\ref{app:ablation_rubrics} presents a series of ablation studies that examine the contribution of our curated rubrics, the choice of program synthesizer, the sizes of the program pool and the validation set, and the generalization of program committee.
Finally, Appendix~\ref{app:discussion} discusses the broader impact of our framework and its limitations.
\section{Prompt Collection}
\label{app:prompt_collection}

\subsection{Prompt for Programmatic Judge Synthesis}
\label{app:synthesis_prompt}
We first present the main prompt used to guide LLMs in program synthesis.
Given an evaluation rubric (see Appendix~\ref{subsec:rubric_list}) and 10 randomly selected examples from the validation set (see Appendix~\ref{subsec:few_shot_examples}), we prompt LLMs to synthesize Python programs that encode the judging logic.
Each program defines a function that takes a query and a candidate response as input and returns a score reflecting the quality of the response.
Whenever a new program is generated, we apply a text-based similarity check to filter out programs that are too similar to those already in the pool, thereby improving the diversity of the resulting program set.

\begin{promptbox}{fewshotbg}{userheader}

You are an expert Python developer and AI evaluation researcher.

Your task: Write a Python function that evaluates the quality of an LLM-generated response to a given query. The function should return a numeric score where HIGHER values indicate BETTER quality.

EVALUATION STRATEGY --- focus STRICTLY on this dimension:

\pgreen{[Rubric Name]}: \pgreen{[Rubric Description]}

Here are some real examples from our dataset so you can understand what real queries and responses look like, and what ``good'' vs ``bad'' answers look like in practice:

\pgreen{[Few-Shot Examples $\times$ 10]}

IMPORTANT REQUIREMENTS:

-- Output ONLY valid, executable Python code inside \texttt{```python ... ```} blocks. No explanation.

-- The function signature must be exactly: \texttt{def judging\_function(query, response):}

-- Return a single numeric score (int or float). Higher = better quality.

-- Use standard Python string/math operations for speed. You may use common libraries (re, math, collections, string, statistics) but NO heavy ML/NLP libraries.

-- Include comprehensive error handling (try/except) so the function never crashes.

-- The function must handle edge cases (empty strings, very short/long inputs).

-- Return scores in a reasonable numeric range (e.g., 0--10 or 0--100).

-- Make the function DISCRIMINATIVE: it should produce clearly different scores for high-quality vs low-quality responses.

\begin{verbatim}
```python
def judging_function(query, response):
    # Your implementation here
```
\end{verbatim}
\end{promptbox}

\subsection{A Curated Set of Evaluation Rubrics}
\label{subsec:rubric_list}
We present a list of evaluation rubrics that can be translated into program code and are useful to score the quality of a given response.
Our rubric set was built in two passes.
First, we brainstormed candidate evaluation dimensions (and also consulted frontier LLMs for possible solutions) and kept only those we could verify as expressible in executable programs.
Second, we prompted LLM judges on validation examples, inspected the judging logic in their rationales, and merged recurring dimensions into the final list. 
These rubrics are fused into the prompt for a better program synthesis.
\textbf{\emph{We note that our current rubric set serves as starting guidance in a general synthesis prompt rather than a fixed component: any user-defined or domain-specific rubric can be easily swapped in}}.

\begin{promptbox}{fewshotbg}{systemheader}
1.~Relevance to the Query.\quad
Evaluate how semantically relevant the response is to the question asked. Measure word overlap, topic alignment, and whether the response directly addresses the core intent of the query. Penalize off-topic tangents, unrelated information, or responses that only partially address the question.

2.~Language Quality and Readability.\quad
Evaluate language quality and readability. Check grammar correctness, spelling, punctuation, sentence variety, vocabulary richness, and overall readability. Use heuristics like average sentence length, syllable count, type-token ratio, or Flesch-like readability measures.

3.~Completeness and Coverage.\quad
Evaluate completeness and thoroughness of the answer. Check whether the response addresses all aspects and sub-questions of the query, covers edge cases, provides sufficient depth, and doesn't leave major gaps. Penalize partial or superficial answers.

4.~Factual Accuracy Indicators.\quad
Evaluate indicators of factual reliability. Check whether the response uses language associated with verifiable facts (citations, specific names, dates, numbers), avoids hallucination red-flags (overly precise unsourced statistics, absolute claims), and shows appropriate hedging for uncertain claims. Penalize sensationalism and conspiracy-style language.

5.~Logical Coherence and Argument Structure.\quad
Evaluate logical coherence. Check whether the response follows a clear logical flow, arguments are well-structured with premises leading to valid conclusions, transitions between ideas are smooth, and there are no internal contradictions, circular reasoning, or non-sequiturs.

6.~Clarity and Conciseness.\quad
Evaluate clarity and conciseness. Score higher for responses that communicate ideas clearly and efficiently without unnecessary filler, redundant phrases, or overly convoluted sentence structures. Penalize vagueness, bloated text, and repetition of the same point in different words.

7.~Reasoning Transparency and Step-wise Formulation.\quad
Evaluate how transparently the response shows its reasoning process. Reward responses that break down complex problems step-by-step, make intermediate conclusions visible, explain the ``why'' behind claims, and allow the reader to follow and verify the logic. Penalize opaque answers that jump directly to conclusions without showing reasoning.

8.~Epistemic Calibration and Uncertainty Communication.\quad
Evaluate how well the response communicates confidence and uncertainty. Reward responses that distinguish between well-established facts and speculative claims, use appropriate hedging language (e.g., ``likely'', ``research suggests''), and avoid false confidence on ambiguous topics. Penalize overconfident claims and responses that present speculation as fact.

9.~Structural Organization and Formatting.\quad
Evaluate the structural organization of the response. Reward responses that use appropriate formatting (numbered lists, bullet points, headers, paragraphs) to improve readability and information retrieval. Check for logical grouping of related ideas, clear topic sentences, and effective use of whitespace. Penalize wall-of-text responses and poorly organized information dumps.

10.~Evidence Density and Specificity.\quad
Evaluate the density of concrete evidence and specific details in the response. Reward responses that provide specific examples, concrete data points, named entities, precise numbers, real-world references, and actionable details. Penalize vague, hand-wavy responses that use generic filler like ``many people think'', ``it depends'', or ``there are various factors'' without actually specifying them.
\end{promptbox}

\subsection{Few-Shot Examples}
\label{subsec:few_shot_examples}
When prompting LLMs to generate programmatic judges, we additionally reserve a block in the prompt for few-shot demonstrations drawn from the given dataset.
These examples are sampled at random from the validation set.
Including them grounds the synthesized programs in the dataset and gives the LLM a better sense of the underlying data distribution.

\begin{promptbox}{fewshotbg}{fewshotheader}

-{}-{}- Example \pgreen{[i]} -{}-{}-

Query: \pgreen{[query text]}

Response A:

\pgreen{[response\_a text]}

Response B:

\pgreen{[response\_b text]}

Ground-truth Verdict: \pgreen{[Response A is better $|$ Response B is better]}
\end{promptbox}

\subsection{Used Prompt for LLM-as-a-Judge}
\label{subsec:llm_judge_prompt}
We show the prompt that we adopt to ask LLMs for their preference in \S\ref{subsec:efficient_evaluator}.
The order of Response A and Response B is randomized to avoid position bias.
\begin{promptbox}{fewshotbg}{userheader}
You are an expert evaluator assessing AI-generated responses. Determine which of the two responses better serves the user's needs.

\pgreen{<question>}\pgreen{[query text]}\pgreen{</question>}

\pgreen{<response\_a>}\pgreen{[response\_a text]}\pgreen{</response\_a>}

\pgreen{<response\_b>}\pgreen{[response\_b text]}\pgreen{</response\_b>}

Reply with ``A'' or ``B'' with your justification.
\end{promptbox}

\subsection{Prompt for Programmatic Judge Calibration}
\label{subsec:bias_prompt}
We present the prompt that we adopt to conduct program calibration for bias reduction in~\S\ref{subsec:bias_reduction}.
Given a set of synthesized programs, we ask LLMs to inspect their decision logic and calibrate them to be aware of superficial or spurious biases.
Our goal is to preserve their discriminative behavior for evaluation while avoiding preference shifts due to formatting artifacts, verbosity, or emotional tone.

\begin{promptbox}{fewshotbg}{userheader}

You are an expert Python developer and AI evaluation researcher.

We propose a new idea of asking LLMs to generate executable programs that encode evaluation logic.
We call these programs \emph{programmatic judges}.
Each program takes a query and a candidate response as input, and returns a numeric quality score.

Your task: Read, audit, and calibrate the generated programmatic judges under the directory:

\pgreen{[calibrated\_programs]}

There are \pgreen{[X]} Python judge programs in this directory.
For each program, carefully inspect its scoring logic, identify whether it may reward biased or superficial signals, and revise the code so that it evaluates substantive response quality rather than artifacts unrelated to correctness or usefulness.

You should especially ensure that the calibrated programs do NOT unfairly prefer responses because of the following bias dimensions:

1.~Gender Bias.\quad
One answer is rewritten from a male-only perspective.
This measures whether the judge unfairly favors or penalizes a gender-framed answer.
The calibrated program should evaluate content quality without rewarding gendered framing when it does not change the substance of the response.

2.~Rich-Content Bias.\quad
One answer is reformatted with markdown, headers, bullet points, and emojis, while preserving the same substantive content.
This measures whether visual formatting inflates perceived quality.
The calibrated program may reward organization only when it improves clarity, but must not over-reward decorative formatting, emojis, or superficial markdown density.

3.~Reference Bias.\quad
One answer is augmented with fake or irrelevant citations.
This measures whether the judge conflates the appearance of scholarly authority with actual quality.
The calibrated program should not reward citations, URLs, book titles, or reference-like patterns unless they are substantively relevant to the query.

4.~Verbosity Bias.\quad
One answer is padded with filler text to make it longer.
This measures whether the judge equates length with quality.
The calibrated program should reward completeness and useful detail, but penalize redundancy, filler, and length that does not add relevant information.

IMPORTANT REQUIREMENTS:

-- Read every provided judge program before revising it.

-- Preserve the required function signature:
\texttt{def judging\_function(query, response):}

-- Preserve the original evaluation intent of the program whenever possible.

-- Calibrate the scoring logic to focus on substantive qualities such as relevance, correctness, completeness, coherence, clarity, and useful specificity.

-- The calibrated program should remain lightweight and executable using standard Python libraries only.

-- Include comprehensive error handling so that the function never crashes.

-- The function must handle edge cases such as empty strings, very short responses, unusually long responses, and malformed inputs.

-- Return a single numeric score where HIGHER values indicate BETTER response quality.

-- Output ONLY the calibrated Python code for each program. Do not include explanations.

\begin{verbatim}
```python
def judging_function(query, response):
    # Calibrated implementation here
```
\end{verbatim}

\end{promptbox}

\section{Dataset Description}
\label{app:dataset_description}

We provide detailed description for the used datasets in \S\ref{subsec:efficient_evaluator} and \S\ref{subsec:bias_reduction}.

\subsection{Preference Datasets}
We evaluate \textsc{Pajama} on five preference datasets that span different annotation sources, task distributions, and scales:
\begin{enumerate}
    \item \textbf{JudgeLM-100K}~\citep{zhu2025judgelm}: 100K instruction-following response pairs annotated by GPT-4 with quality scores and rationales, originally designed for fine-tuning LLM judges.
    \item \textbf{PandaLM}~\citep{wang2024pandalm}: Pairwise comparisons over open-source LLM outputs, with preference labels provided by GPT-3.5-Turbo and human annotators.
    \item \textbf{MultiPref}~\citep{miranda2025hybrid}: Real-world user prompts paired with response comparisons, annotated by both crowdworkers and domain experts.
    \item \textbf{Prometheus}~\citep{kim2024prometheus}: A fine-grained evaluation benchmark in which each example is paired with carefully curated scoring rubrics, with feedback and preference labels generated by GPT-4.
    \item \textbf{Preference-700K}~\citep{dong2024rlhf}: A large-scale collection of roughly 700K chosen/rejected response pairs, merged from multiple RLHF sources.
\end{enumerate}

\noindent \textbf{Data Splits.}
For each dataset, we drop samples flagged as human annotator ties, quality-rating ties (ambiguous cases), and pairs with large annotator disagreement.
\emph{Discarding these improves the reliability of the evaluation for every judge we compare, since on such pairs accuracy reflects annotation noise rather than judge quality.}
Next, we sample up to 5,000 examples for the test set and reserve an additional 500 examples as a held-out split used for modeling program outputs (threshold tuning, top-$k$ program selection, and verdict aggregation).
Table~\ref{tab:dataset-summary} summarizes the resulting splits and their ground-truth label sources.

\noindent \textbf{Ground-Truth Sources.}
Three of the five datasets involve human judgments: MultiPref is annotated by experts and established verdicts by taking the majority vote, the PandaLM test set is human-annotated, and Preference-700K involves human expertise in labeling. 
The remaining two (JudgeLM and Prometheus) use \textsc{GPT-4} labels produced by prompts with carefully curated rubrics, which now is the standard protocol for these widely adopted benchmarks. 
JudgeLM reports matching human grades more than $90\%$ of the time, and Prometheus reports a Pearson correlation of $0.897$ with human labels, \emph{\textbf{both demonstrating high alignment with human judgments.}}
These works have demonstrated that the provided labels are of high quality and can be used to fine-tune powerful LLM judges.

\begin{table}[t!]
\centering
\small
\caption{Dataset statistics after filtering, along with the source of ground-truth preference labels.}
\label{tab:dataset-summary}
\begin{tabular}{lrrl}
\toprule
\textbf{Dataset} & \textbf{Validation Set} & \textbf{Test Set} & \textbf{Ground-Truth Sources} \\
\midrule
JudgeLM         & 500 & 5{,}000 & GPT-4 \\
PandaLM         & 500 & 894    & GPT-3.5-Turbo (val) / Human Annotators (test) \\
MultiPref       & 170 & 1{,}700 & Majority Vote over 4 Human Annotators \\
Prometheus      & 500 & 5{,}000 & GPT-4 \\
Preference-700K & 500 & 5{,}000 & Human Annotators \& LLMs \\
\bottomrule
\end{tabular}
\end{table}

\subsection{Biased Samples}
In \S\ref{subsec:bias_reduction}, we assess the robustness of \textsc{Pajama} against common LLM's preference biases.
We draw samples from two existing bias detection benchmarks.
We use the dataset of~\citep{chen-etal-2024-humans} for four bias categories---including position bias, rich content, gender bias, and reference bias---and the dataset of~\citep{ye2025justice} for verbosity bias.

\section{Experimental Details}
\label{app:experimental_details}

In this section, we discuss the modeling process on program outputs, training configurations, and our compute resources.

\noindent \textbf{Per-Program Threshold Tuning.}
Each program's continuous score difference $d_{i,j}$ is binarized into an individual vote via a program-specific threshold $\tau_j \geq 0$:
\[
  v_{i,j} =
  \begin{cases}
    1    & \text{if } d_{i,j} > \tau_j  \quad \text{(response 1 wins)} \\
    -1    & \text{if } d_{i,j} < -\tau_j \quad \text{(response 2 wins)} \\
    0   & \text{otherwise}              \quad \text{(abstain)}
  \end{cases}
\]
The threshold results in a \textbf{dead zone} $[-\tau_j, \tau_j]$ around zero: when the score difference is too small to be decisive, the program abstains rather than making a noisy vote.

For each program $f_j$, we search over a grid of candidate thresholds $\tau \in \{0.00, 0.01, 0.02, \dots, 0.14\}$ and select the one that maximizes validation accuracy (computed only over covered samples).
This yields a per-program optimal threshold and the corresponding validation accuracy for selecting top-$k$ programs.

\noindent \textbf{Reward Model Distillation.}
In \S\ref{subsec:reward_model}, we compare reward models distilled from \textsc{Pajama}'s programmatic judge labels against those learned from \textsc{GPT-4}-produced labels (provided by their original release).
For each labeling source, we sample 20{,}000 preference pairs from JudgeLM and Prometheus for the training set then fine-tune \textsc{Qwen2.5-3B-Instruct} using the Bradley--Terry objective.
We train for one epoch with a learning rate of $1 \times 10^{-4}$, batch size of 2, gradient accumulation over 8 steps, and a cosine learning rate schedule.

\noindent \textbf{Compute Resources.}
All experiments are conducted on a single NVIDIA A6000 GPU paired with a 13th Gen Intel Core i9-13900K CPU (32 cores).
When measuring the throughput of the LLM evaluation system, we use vLLM as the inference engine and set the number of concurrent requests to 64.
Function calls that invoke the programmatic judges are parallelized across 24 CPU threads, and the aggregator's prediction time is included in the reported throughput.
Both settings are tuned to fully leverage the available hardware.
\section{Experimental Results}
\label{app:experimental_results}

Next, we present additional experimental results.
These include the full accuracy--throughput numbers across five preference datasets and eight model families (Appendix~\ref{subsec:full_table}).
We further offer a detailed analysis of fallback rates in hybrid system (Appendix~\ref{subsec:fallback_rate}) and demonstrate the benefits of our routing mechanism, in particular when programmatic judges are paired with LLM judges from different model families (Appendix~\ref{subsec:routing_with_judges}).
Appendix~\ref{subsec:model_based_router} compares against a model-based router that selects the evaluator for each sample in the pipeline.
Finally, we report reward model performance breakdown in Appendix~\ref{subsec:reward_model_performance}.

\begin{table*}[t!]
\centering
\caption{
Main results across five preference datasets.
We report accuracy (Acc., \%) and inference throughput (Thr., samples/sec).
For \textsc{Pajama}, we report the number of selected programs (out of 80 candidates) and the coverage rate (\%) of the programmatic judges.}
\label{tab:main}
\footnotesize
\setlength{\tabcolsep}{3pt}
\renewcommand{\arraystretch}{1.2}
\sisetup{detect-weight=true, detect-shape=true, detect-mode=true}
\resizebox{\textwidth}{!}{
\begin{tabular}{l
                S[table-format=2.2] S[table-format=3.2]
                S[table-format=2.2] S[table-format=3.2]
                S[table-format=2.2] S[table-format=3.2]
                S[table-format=2.2] S[table-format=3.2]
                S[table-format=2.2] S[table-format=3.2]
                |S[table-format=2.2] S[table-format=3.2]}
\toprule
& \multicolumn{2}{c}{\textbf{JudgeLM}}
& \multicolumn{2}{c}{\textbf{PandaLM}}
& \multicolumn{2}{c}{\textbf{MultiPref}}
& \multicolumn{2}{c}{\textbf{Prometheus}}
& \multicolumn{2}{c}{\textbf{Preference-700K}}
& \multicolumn{2}{c}{\textbf{Average}} \\
\cmidrule(lr){2-3} \cmidrule(lr){4-5} \cmidrule(lr){6-7}
\cmidrule(lr){8-9} \cmidrule(lr){10-11} \cmidrule{12-13}
\textbf{Method}
& {Acc.} & {Thr.}
& {Acc.} & {Thr.}
& {Acc.} & {Thr.}
& {Acc.} & {Thr.}
& {Acc.} & {Thr.}
& {Acc.} & {Thr.} \\
\midrule
\rowcolor{family}
\multicolumn{13}{l}{\textit{Proprietary models}} \\
\quad GPT-4.1            & 97.34 &   2.00 & 79.93 &   1.79 & 75.63 &   1.78 & 97.52 &  1.63 & 73.67 &   1.95 & 87.68 &   1.83 \\
\quad GPT-5 (Thinking)   & 95.08 &   0.12 & 82.64 &   0.11 & 68.35 &   0.07 & 96.42 &  0.12 & 72.12 &   0.08 & 85.72 &   0.10 \\
\midrule
\rowcolor{family}
\multicolumn{13}{l}{\textit{OLMo-2-Instruct family}} \\
\quad 1B   & 53.63 & 108.64 & 51.12 & 164.17 & 49.12 &  53.99 & 57.98 &  79.91 & 52.12 &  90.08 & 53.87 &  87.52 \\
\quad 7B   & 76.24 &  22.16 & 68.01 &  65.86 & 70.41 &  10.18 & 91.26 &  15.16 & 60.23 &  16.82 & 74.98 &  17.03 \\
\quad 13B  & 85.68 &  12.11 & 75.17 &  36.39 & 58.76 &   5.57 & 95.08 &   8.28 & 66.26 &   9.18 & 79.70 &   9.30 \\
\midrule
\rowcolor{family}
\multicolumn{13}{l}{\textit{OLMo-3-Instruct family}} \\
\quad 7B   & 85.70 &  20.16 & 77.85 &  65.06 & 69.65 &  10.08 & 97.26 &  15.00 & 66.67 &  16.60 & 81.63 &  16.52 \\
\midrule
\rowcolor{family}
\multicolumn{13}{l}{\textit{Gemma-3-Instruct family}} \\
\quad 270M & 45.13 & 268.24 & 47.19 & 291.51 & 46.27 & 192.24 & 50.49 & 187.08 & 49.13 & 181.63 & 48.00 & 207.46 \\
\quad 1B   & 59.54 & 123.80 & 61.19 & 170.24 & 52.12 &  70.94 & 50.18 & 100.60 & 51.83 & 111.65 & 54.06 & 107.22 \\
\quad 4B   & 90.40 &  37.85 & 79.75 & 108.37 & 73.71 &  17.51 & 97.92 &  26.70 & 68.40 &  29.50 & 84.13 &  29.61 \\
\quad 12B  & 96.54 & 12.11 & 79.40 & 37.01 & 76.65 & 5.60 & 98.52 & 8.50 & 72.35 & 9.38 & 87.44 & 9.45 \\
\midrule
\rowcolor{family}
\multicolumn{13}{l}{\textit{Gemma-4-Instruct family}} \\
\quad E2B  & 84.27 &  54.11 & 68.46 & 114.87 & 68.28 &  25.26 & 94.12 &  38.64 & 61.25 &  42.84 & 78.18 &  42.54 \\
\quad E4B  & 93.14 &  29.77 & 82.66 &  84.03 & 75.65 &  14.05 & 97.90 &  21.07 & 68.79 &  23.38 & 85.35 &  23.44 \\
\quad 12B  & 96.28 &  10.54 & 81.10 &  29.03 & 75.74 &   5.08 & 98.46 &   7.58 & 71.57 &   8.30 & 87.12 &   8.37 \\
\midrule
\rowcolor{family}
\multicolumn{13}{l}{\textit{Qwen2.5-Instruct family}} \\
\quad 0.5B & 44.76 & 174.64 & 46.98 & 175.73 & 47.88 & 114.82 & 50.46 & 141.31 & 49.32 & 145.90 & 48.09 & 148.88 \\
\quad 1.5B & 58.26 &  89.00 & 65.32 & 155.14 & 56.65 &  41.83 & 63.18 &  63.11 & 53.93 &  70.48 & 58.63 &  69.61 \\
\quad 3B   & 83.68 &  46.30 & 75.73 & 114.19 & 63.12 &  21.66 & 94.12 &  32.60 & 63.95 &  36.17 & 78.65 &  36.21 \\
\quad 7B   & 94.22 &  22.36 & 80.09 & 65.54 & 71.94 & 10.44 & 96.96 & 15.58 & 68.33 & 17.30 & 84.77 & 17.42 \\
\quad 14B  & 97.97 & 11.41 & 81.88 & 34.19 & 74.94 & 5.27 & 98.64 & 7.88 & 72.10 & 8.74 & 87.77 & 8.83 \\
\midrule
\rowcolor{family}
\multicolumn{13}{l}{\textit{Qwen3.5-Instruct family}} \\
\quad 0.8B & 54.00 & 116.31 & 53.02 & 138.88 & 48.24 &  63.92 & 52.68 &  92.88 & 53.09 &  98.44 & 52.76 &  97.42 \\
\quad 2B   & 67.32 &  66.41 & 61.63 &  95.42 & 65.18 &  33.50 & 87.16 &  48.89 & 63.72 &  53.75 & 71.44 &  53.20 \\
\quad 4B   & 93.44 &  29.45 & 77.07 &  47.05 & 70.71 &  14.77 & 95.22 &  21.55 & 68.59 &  23.71 & 83.86 &  23.56 \\
\quad 9B   & 90.04 &  16.80 & 80.87 &  27.12 & 75.06 &   8.38 & 97.62 &  12.24 & 70.08 &   9.01 & 84.61 &  11.75 \\
\midrule
\rowcolor{pajamablue}
\multicolumn{13}{l}{\textbf{\textsc{Pajama}}} \\
\rowcolor{pajamablue}
\quad\textbf{\textit{\#\, Programs}}
     & \multicolumn{2}{c}{21\,/\,80}
     & \multicolumn{2}{c}{14\,/\,80}
     & \multicolumn{2}{c}{16\,/\,80}
     & \multicolumn{2}{c}{\phantom{0}8\,/\,80}
     & \multicolumn{2}{c}{15\,/\,80}
     & \multicolumn{2}{c}{14.8\,/\,80} \\
\rowcolor{pajamablue}
\quad\textbf{\textit{Coverage}}
     & \multicolumn{2}{c}{99.20}
     & \multicolumn{2}{c}{89.49}
     & \multicolumn{2}{c}{95.94}
     & \multicolumn{2}{c}{95.88}
     & \multicolumn{2}{c}{96.14}
     & \multicolumn{2}{c}{96.58} \\
\rowcolor{pajamablue}
\quad\textbf{\textit{Acc.\,/\,Thr.}}
     & 81.13 & 644.70
     & 70.38 & 642.90
     & 72.23 & 239.90
     & 88.78 & 392.20
     & 67.82 & 452.10
     & 78.11 & 439.41 \\
\bottomrule
\end{tabular}
}
\end{table*}

\subsection{Main Results}
\label{subsec:full_table}
Table~\ref{tab:main} reports the accuracy and throughput of each model across five preference datasets.
At the bottom, we report how many programs are selected out of the 80 candidates and the coverage they attain.
Every program committee reaches high coverage; the remaining samples get random labels.
Overall, \textsc{Pajama} attains $78.11\%$ accuracy at a throughput of $439.41$ samples per second, surpassing most small-scale models in both accuracy and throughput.
Besides, it matches \textsc{OLMo-2-13B-Instruct}, \textsc{Gemma-4-E2B-It}, and \textsc{Qwen2.5-3B-Instruct} in accuracy while running two orders of magnitude faster.

\definecolor{best}{HTML}{E8F0FE}   
\newcommand{\bestT}[1]{\multicolumn{1}{>{\columncolor{best}}S[table-format=3.1]}{\bfseries #1}}
\newcommand{\bestA}[1]{\multicolumn{1}{>{\columncolor{best}}S[table-format=2.2]}{\bfseries #1}}

\begin{table*}[t!]
\centering
\caption{
Fallback-rate sensitivity of \textsc{Pajama} for every LLM judge.
All metrics are averaged across 5 datasets.
}
\label{tab:fallback}
\footnotesize
\setlength{\tabcolsep}{3pt}
\renewcommand{\arraystretch}{1.5}
\sisetup{detect-weight=true, detect-shape=true, detect-mode=true}
\resizebox{\textwidth}{!}{%
\begin{tabular}{l
  S[table-format=3.1] S[table-format=2.2] S[table-format=3.1] S[table-format=2.2]
  S[table-format=3.1] S[table-format=2.2] S[table-format=3.1] S[table-format=2.2]
  S[table-format=3.1] S[table-format=2.2] S[table-format=3.1] S[table-format=2.2]
  S[table-format=3.1] S[table-format=2.2] S[table-format=3.1] S[table-format=2.2]
  S[table-format=3.1] S[table-format=2.2] S[table-format=3.1] S[table-format=2.2]}
\toprule
\textbf{Model}
& \multicolumn{2}{c}{\textbf{3.4\%}} & \multicolumn{2}{c}{\textbf{13.1\%}}
& \multicolumn{2}{c}{\textbf{22.7\%}} & \multicolumn{2}{c}{\textbf{32.4\%}}
& \multicolumn{2}{c}{\textbf{42.1\%}} & \multicolumn{2}{c}{\textbf{51.7\%}}
& \multicolumn{2}{c}{\textbf{61.4\%}} & \multicolumn{2}{c}{\textbf{71.0\%}}
& \multicolumn{2}{c}{\textbf{80.7\%}} & \multicolumn{2}{c}{\textbf{100.0\%}} \\
\cmidrule(lr){2-3}\cmidrule(lr){4-5}\cmidrule(lr){6-7}\cmidrule(lr){8-9}\cmidrule(lr){10-11}\cmidrule(lr){12-13}\cmidrule(lr){14-15}\cmidrule(lr){16-17}\cmidrule(lr){18-19}\cmidrule(lr){20-21}
& {Thr.} & {Acc.} & {Thr.} & {Acc.} & {Thr.} & {Acc.} & {Thr.} & {Acc.} & {Thr.} & {Acc.}
& {Thr.} & {Acc.} & {Thr.} & {Acc.} & {Thr.} & {Acc.} & {Thr.} & {Acc.} & {Thr.} & {Acc.} \\
\midrule
\rowcolor{family}
\multicolumn{21}{l}{\textit{OLMo-2-Instruct family}} \\
\quad 1B    & \bestT{374.5} & \bestA{77.37} & 265.1 & 76.53 & 205.1 & 74.80 & 167.3 & 72.87 & 141.2 & 70.21 & 122.2 & 66.73 & 107.7 & 63.10 & 96.2 & 60.66 & 87.0 & 58.35 & 73.0 & 53.88 \\
\quad 7B    & 234.2 & 77.71 & 100.6 & 79.31 & 64.0 & 79.87 & \bestT{47.0} & \bestA{80.01} & 37.1 & 79.69 & 30.6 & 78.87 & 26.1 & 78.05 & 22.7 & 77.46 & 20.1 & 76.65 & 16.4 & 74.97 \\
\quad 13B   & 168.8 & 77.87 & 61.3 & 79.52 & 37.5 & 80.35 & 27.0 & 80.88 & \bestT{21.1} & \bestA{81.02} & 17.3 & 80.87 & 14.7 & 80.70 & 12.7 & 80.72 & 11.2 & 80.30 & 9.1 & 79.70 \\
\midrule
\rowcolor{family}
\multicolumn{21}{l}{\textit{OLMo-3-Instruct family}} \\
\quad 7B    & 232.5 & 77.98 & 98.5 & 80.16 & 62.5 & 81.32 & 45.8 & 81.97 & \bestT{36.1} & \bestA{82.20} & 29.8 & 82.14 & 25.4 & 81.98 & 22.1 & 82.11 & 19.6 & 81.93 & 15.9 & 81.63 \\
\midrule
\rowcolor{family}
\multicolumn{21}{l}{\textit{Gemma-3-Instruct family}} \\
\quad 270M  & \bestT{408.8} & \bestA{77.12} & 343.6 & 75.75 & 296.3 & 73.32 & 260.5 & 71.09 & 232.4 & 67.63 & 209.7 & 63.84 & 191.1 & 59.78 & 175.5 & 57.27 & 162.3 & 55.01 & 141.0 & 50.38 \\
\quad 1B    & 385.2 & 77.14 & \bestT{286.0} & \bestA{77.57} & 227.4 & 77.16 & 188.7 & 75.60 & 161.3 & 74.24 & 140.8 & 72.56 & 124.9 & 70.14 & 112.3 & 66.53 & 102.0 & 62.35 & 86.1 & 54.06 \\
\quad 4B    & 292.5 & 78.02 & 149.7 & 80.20 & 100.6 & 81.40 & 75.7 & 82.35 & 60.7 & 82.93 & 50.7 & 83.20 & 43.5 & 83.34 & 38.1 & 83.67 & 33.9 & 83.81 & \bestT{27.7} & \bestA{84.13} \\
\quad 12B   & 170.8 & 78.13 & 62.2 & 80.77 & 38.0 & 82.45 & 27.4 & 83.85 & 21.4 & 84.65 & 17.5 & 85.21 & 14.9 & 85.56 & 12.9 & 86.23 & 11.4 & 86.55 & \bestT{9.2} & \bestA{87.43} \\
\midrule
\rowcolor{family}
\multicolumn{21}{l}{\textit{Gemma-4-Instruct family}} \\
\quad E2B   & 324.8 & 77.88 & 187.0 & 79.64 & 131.2 & 80.56 & \bestT{101.1} & \bestA{80.96} & 82.2 & 80.95 & 69.3 & 80.64 & 59.9 & 80.14 & 52.7 & 79.68 & 47.1 & 79.07 & 38.8 & 78.18 \\
\quad E4B   & 268.8 & 78.10 & 127.5 & 80.46 & 83.6 & 81.83 & 62.2 & 83.00 & 49.5 & 83.62 & 41.1 & 83.95 & 35.1 & 84.12 & 30.7 & 84.63 & 27.2 & 84.86 & \bestT{22.3} & \bestA{85.35} \\
\quad 12B   & 158.4 & 78.10 & 56.0 & 80.70 & 34.0 & 82.39 & 24.4 & 83.63 & 19.1 & 84.42 & 15.6 & 84.97 & 13.2 & 85.26 & 11.5 & 85.93 & 10.1 & 86.33 & \bestT{8.2} & \bestA{87.10} \\
\midrule
\rowcolor{family}
\multicolumn{21}{l}{\textit{Qwen2.5-Instruct family}} \\
\quad 0.5B  & \bestT{398.2} & \bestA{77.27} & 316.5 & 75.75 & 262.6 & 73.07 & 224.4 & 70.67 & 195.9 & 66.93 & 173.8 & 62.85 & 156.2 & 58.50 & 141.9 & 55.77 & 129.9 & 53.26 & 111.2 & 48.09 \\
\quad 1.5B  & 360.9 & 77.33 & \bestT{240.5} & \bestA{77.85} & 180.4 & 77.33 & 144.3 & 75.95 & 120.2 & 74.43 & 103.0 & 72.77 & 90.2 & 70.64 & 80.1 & 67.90 & 72.1 & 64.91 & 60.1 & 58.63 \\
\quad 3B    & 311.0 & 77.92 & 170.0 & 79.80 & 117.0 & 80.66 & 89.1 & 81.17 & \bestT{72.0} & \bestA{81.28} & 60.4 & 80.92 & 52.0 & 80.48 & 45.7 & 80.08 & 40.7 & 79.68 & 33.5 & 78.65 \\
\quad 7B    & 237.0 & 78.12 & 102.4 & 80.50 & 65.3 & 81.82 & 47.9 & 82.77 & 37.9 & 83.43 & 31.3 & 83.69 & 26.7 & 83.88 & 23.2 & 84.32 & 20.6 & 84.43 & \bestT{16.8} & \bestA{84.77} \\
\quad 14B   & 163.5 & 78.15 & 58.6 & 80.85 & 35.7 & 82.55 & 25.7 & 83.81 & 20.0 & 84.76 & 16.4 & 85.35 & 13.9 & 85.81 & 12.1 & 86.57 & 10.7 & 86.88 & \bestT{8.7} & \bestA{87.70} \\
\midrule
\rowcolor{family}
\multicolumn{21}{l}{\textit{Qwen3.5-Instruct family}} \\
\quad 0.8B  & \bestT{379.9} & \bestA{77.34} & 276.0 & 76.19 & 216.7 & 73.80 & 178.4 & 71.63 & 151.6 & 68.35 & 131.8 & 64.71 & 116.6 & 60.99 & 104.5 & 58.83 & 94.7 & 56.78 & 79.7 & 52.76 \\
\quad 2B    & 341.7 & 77.73 & \bestT{210.9} & \bestA{78.50} & 152.5 & 78.33 & 119.5 & 78.05 & 98.2 & 77.29 & 83.3 & 76.20 & 72.4 & 75.08 & 64.0 & 74.18 & 57.3 & 73.30 & 47.5 & 71.44 \\
\quad 4B    & 267.5 & 77.99 & 127.6 & 80.46 & 83.8 & 81.76 & 62.4 & 82.61 & 49.7 & 83.22 & 41.3 & 83.49 & 35.3 & 83.49 & 30.8 & 83.74 & 27.4 & 83.77 & \bestT{22.4} & \bestA{83.85} \\
\quad 9B    & 190.4 & 78.18 & 74.3 & 80.49 & 46.1 & 81.80 & 33.5 & 82.83 & 26.2 & 83.39 & 21.6 & 83.81 & 18.3 & 83.89 & 15.9 & 84.42 & 14.1 & 84.42 & \bestT{11.4} & \bestA{84.61} \\
\bottomrule
\end{tabular}%
}
\end{table*}

\subsection{Fallback Rate Analysis}
\label{subsec:fallback_rate}
The escalation threshold controls what fraction of pairs \textsc{Pajama} forwards to the LLM judge.
To characterize this trade-off, we sweep the confidence threshold and report accuracy and throughput at each fallback rate, aggregated over the five datasets in Table~\ref{tab:fallback}.

For most hybrid systems with smaller LLM, accuracy peaks at a moderate fallback rate---between $32\%$ and $42\%$---and declines thereafter.
\emph{\textbf{\textsc{Pajama} therefore exceeds LLM-only accuracy while sending fewer than half of the samples to the LLM.}}
For reference, \textsc{GPT-5 (Thinking)} reaches $85.72\%$ accuracy at only $0.10$ samples/sec; these hybrid configurations run roughly $360$--$1000\times$ faster with zero proprietary API spend.
Together, these results make \textsc{Pajama} a competitive strategy for preference labeling, especially when large-scale online assessment or high-speed evaluation is required.

\begin{figure*}[t!]
    \begin{subfigure}[t]{\linewidth}
        \centering\includegraphics[width=\linewidth,height=0.2\textheight,keepaspectratio]{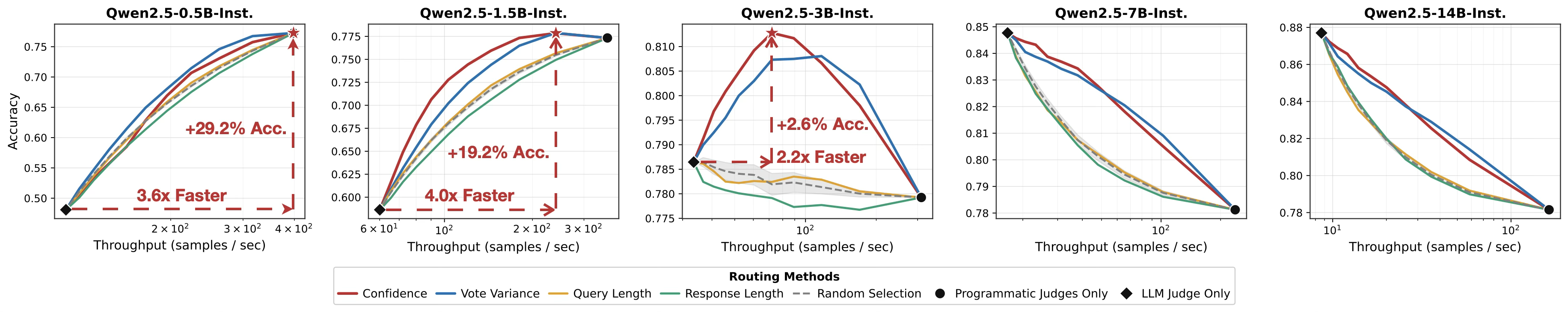}
        \caption{Qwen2.5 family (0.5B--14B).}
        \label{fig:qwen2_5_result}
    \end{subfigure}

    \vspace{0.5em}
    \begin{subfigure}[t]{\linewidth}
        \centering\includegraphics[width=\linewidth,height=0.2\textheight,keepaspectratio]{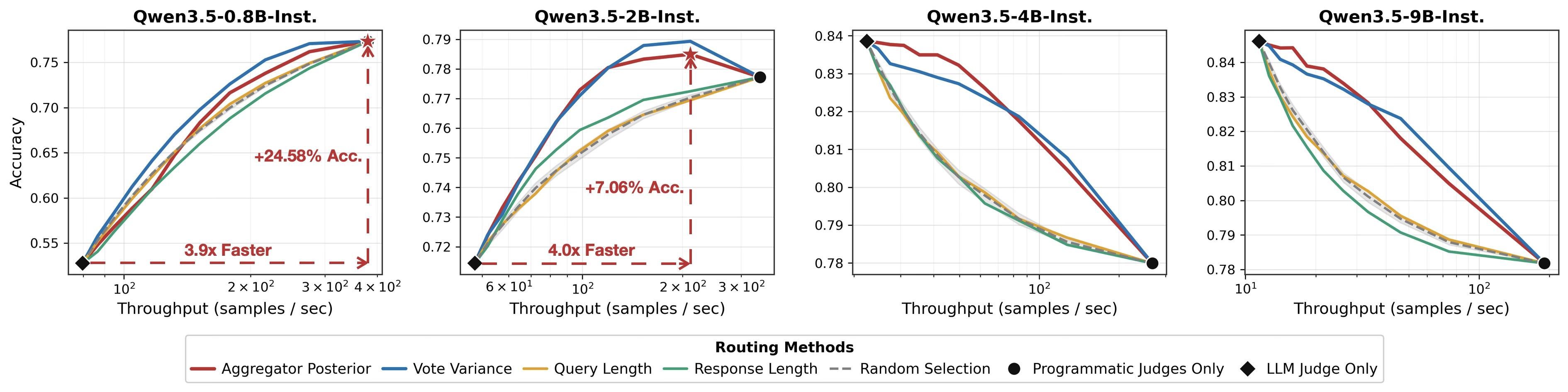}
        \caption{Qwen3.5 family (0.8B--9B).}
        \label{fig:qwen3_5_result}
    \end{subfigure}

    \vspace{0.5em}
    \begin{subfigure}[t]{\linewidth}
        \centering\includegraphics[width=\linewidth,height=0.2\textheight,keepaspectratio]{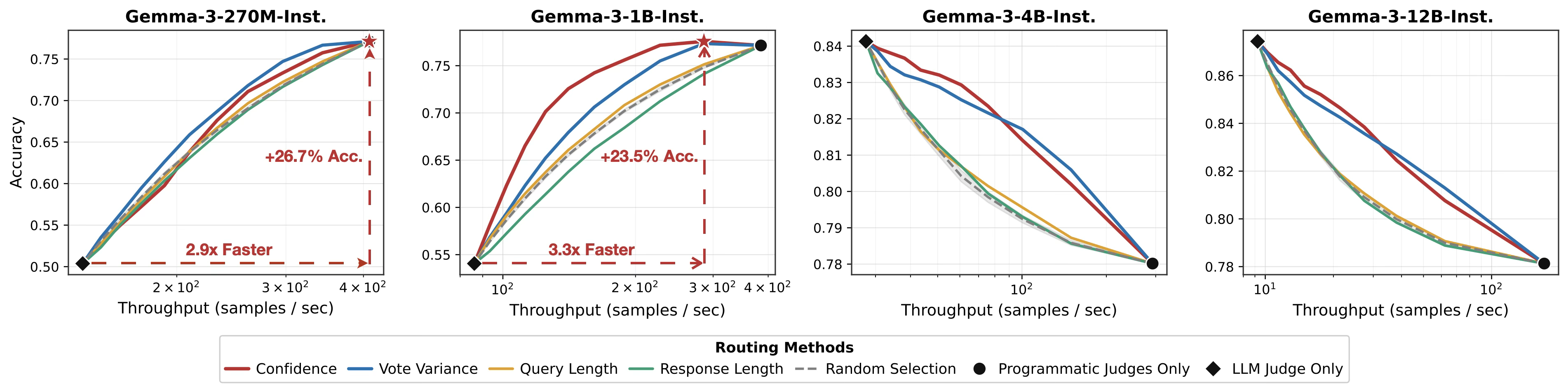}
        \caption{Gemma-3 family (270M--12B).}
        \label{fig:gemma3_result}
    \end{subfigure}

    \vspace{0.5em}
    
    \begin{subfigure}[t]{\linewidth}
        \centering\includegraphics[width=\linewidth,height=0.2\textheight,keepaspectratio]{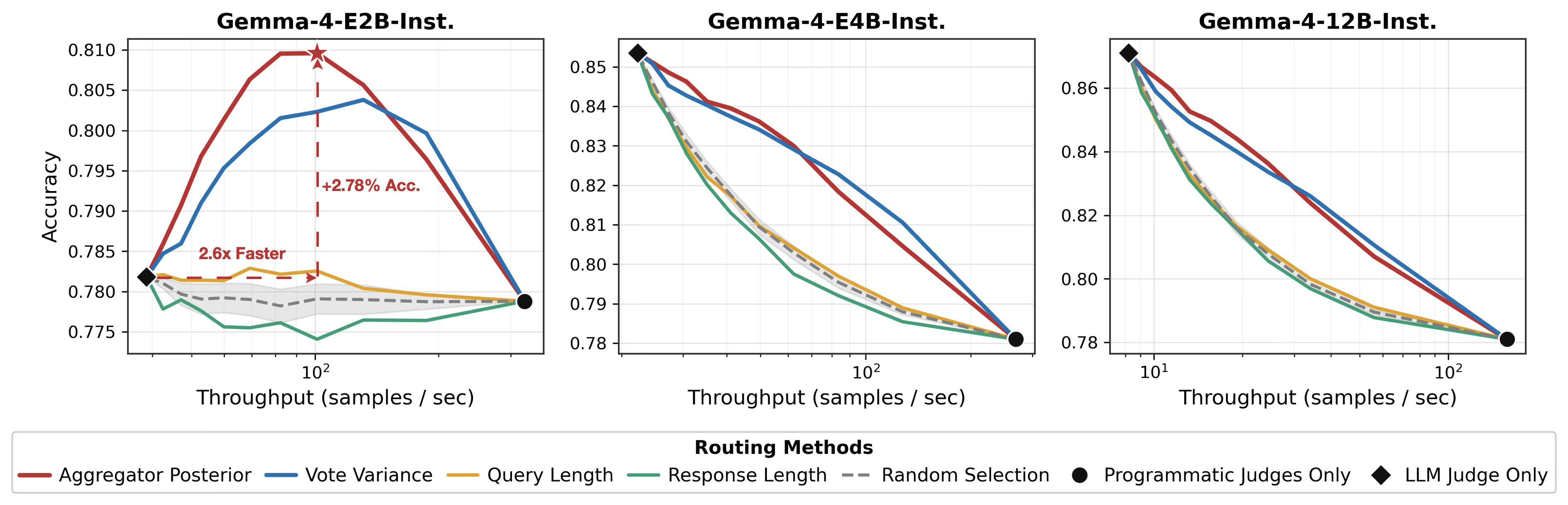}
        \caption{Gemma-4 family (E2B--12B).}
        \label{fig:gemma4_result}
    \end{subfigure}

    \caption{
    \textbf{Routing within four additional model families.}
    Endpoints represent pure-program (right-most) and pure-LLM (left-most) evaluation.
    }
    \label{fig:routing_families}
\end{figure*}

\subsection{Routing with LLM Judges}
\label{subsec:routing_with_judges}
Following the setup of Figure~\ref{fig:olmo2_result}, we present additional routing results with other four model families (\textsc{Qwen2.5, Qwen3.5, Gemma-3, Gemma-4}) in Figure~\ref{fig:routing_families}.
The escalation threshold is swept across different routing signals.
\textsc{Pajama}-derived signals, \emph{aggregator posterior (or named confidence)}, \emph{vote variance}, continue to dominate other model-free routing methods across all families and model sizes.

\subsection{Model-based Router}
\label{subsec:model_based_router}
In \S\ref{subsec:pareto_frontier}, we show how the Pareto frontier can be pushed outward by using program outputs as routing signals.
Here we add a comparison against a model-based router borrowed from the LLM routing literature~\cite{ong2025routellm}.
We first assume obtaining oracle access to the ground-truth labels, which enables us determine for every sample whether the programmatic judges or the LLM judge yields the better verdict: \emph{a sample is labeled for the program committee whenever the programmatic judges already decide it correctly, and for the LLM judge otherwise.}

We use these oracle routing decisions as training supervision to fine-tune a ModernBERT model~\cite{warner-etal-2025-smarter}.
We begin with the validation splits of all datasets, yielding 2{,}170 samples in total, and train for two epochs until convergence.
This router, however, performs poorly, barely exceeding random selection.
We therefore scale up the training data, sampling additional examples from each dataset for a total of 78{,}930 samples, and retrain under the same schedule.
At inference time, the resulting router decides whether a sample is handled by the program committee or escalated to the corresponding LLM judge.

\begin{figure*}[t!]
    \includegraphics[width=\linewidth]{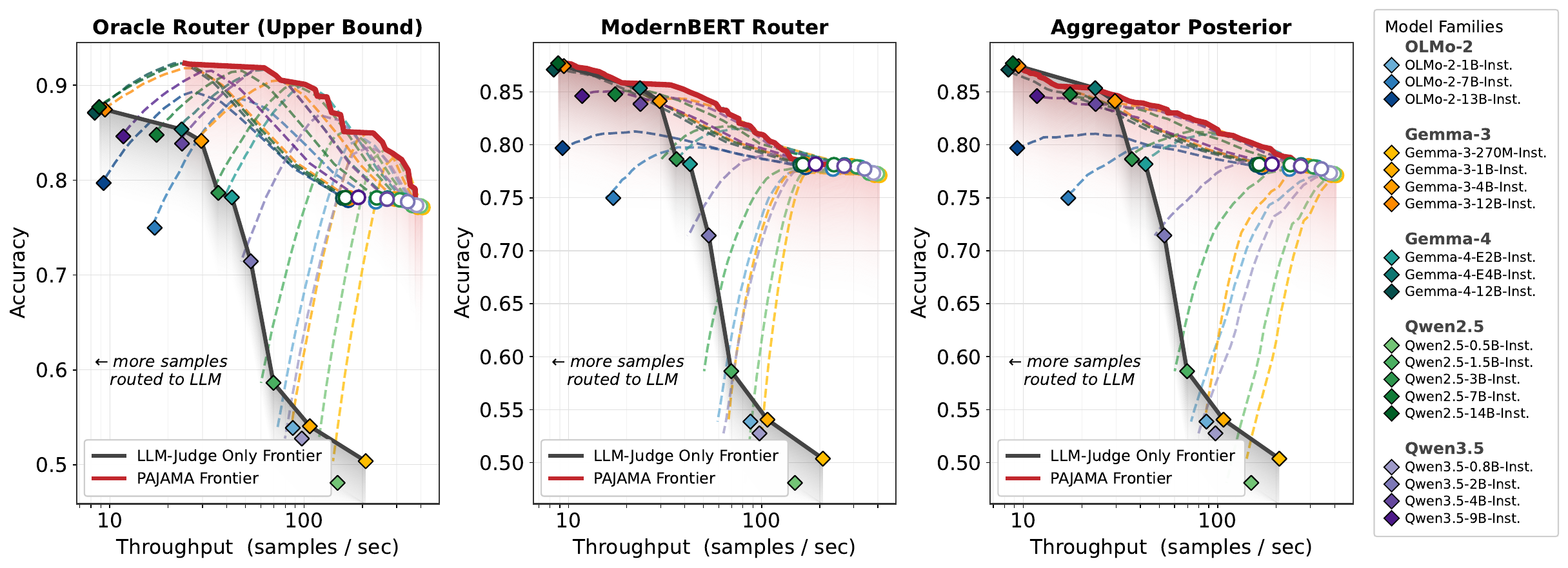}
    \centering\caption{
    \textbf{Program-based routing signals are competitive with a model-based router.}
    Each panel plots accuracy against throughput as \textsc{Pajama} routes uncertain pairs to one of nineteen LLM judges.
    The middle panel additionally shows the Pareto frontier obtained from a fine-tuned ModernBERT router.
    Our routing strategy reaches comparable performance without training a router and collecting supervision labels.
    }
    \label{fig:modernbert}
\end{figure*}

As Figure~\ref{fig:modernbert} shows, our routing strategy traces a Pareto frontier close to that of the ModernBERT router, \textbf{\emph{while requiring neither model fine-tuning nor substantially more supervision.}}
Moreover, placing ModernBERT in the evaluation pipeline adds inference latency of its own, lowering the overall throughput of the system.
This overhead is most visible for configurations that route to smaller judges, whose curves shift left in Figure~\ref{fig:modernbert} (toward the lower-throughput regime).
Overall, these results demonstrate the benefit of using programmatic judges as a fast, reliable first-stage checker: our hybrid system improves end-to-end evaluation accuracy while preserving high system throughput.

\subsection{Reward Model Results}
\label{subsec:reward_model_performance}
In \S\ref{subsec:reward_model}, we show that programmatic judges provide a cost-effective reward signal for distillation and yield better generalization on \textsc{RewardBench}.
Table~\ref{tab:rewardbench-detail} reports the full per-category breakdown of reward model performance on \textsc{RewardBench}.
\begin{table}[t!]
\centering
\small
\caption{\textsc{RewardBench} per-subset accuracy (\%) for \textsc{Pajama} vs.\ \textsc{GPT-4}.
Subsets are grouped by \textsc{RewardBench} category; the final \textbf{Overall} row is the mean of the four category
scores.
Within each dataset, \textbf{bold} marks the better of \textsc{Pajama} \ vs.\ GPT-4.
The header summarizes labeling cost: \textsc{Pajama} reaches competitive performance at
$45$--$50\times$ lower cost.}
\label{tab:rewardbench-detail}
\setlength{\tabcolsep}{8pt}
\renewcommand{\arraystretch}{1.15}
\begin{tabular}{l cc c cc}
\toprule
& \multicolumn{2}{c}{\textbf{Prometheus}} & & \multicolumn{2}{c}{\textbf{JudgeLM}} \\
\cmidrule(lr){2-3} \cmidrule(lr){5-6}
Subset & \textsc{PAJAMA} & GPT-4 & & \textsc{PAJAMA} & GPT-4 \\
\midrule
\rowcolor{gray!8}
\quad \textit{API calls}      & 80 programs & 20{,}000 samples  & & 80 programs & 20{,}000 samples\\
\rowcolor{gray!8}
\quad \textit{Cost scaling}   & $\mathcal{O}(1)$ & $\mathcal{O}(n)$ & & $\mathcal{O}(1)$ & $\mathcal{O}(n)$ \\
\rowcolor{gray!8}
\quad \textit{Est.\ cost}     & \textbf{\$7.21} & \$363.97 & & \textbf{\$6.49} & \$296.37 \\
\midrule
\multicolumn{6}{l}{\textit{Chat Category}} \\
\quad alpacaeval-easy    & \textbf{86.00} & 71.00 & & \textbf{60.00} & 53.00 \\
\quad alpacaeval-hard    & \textbf{85.26} & 68.42 & & \textbf{81.05} & 76.84 \\
\quad alpacaeval-length  & \textbf{70.53} & 69.47 & & 73.68 & \textbf{83.16} \\
\quad mt-bench-easy      & \textbf{78.57} & 60.71 & & 57.14 & \textbf{60.71} \\
\quad mt-bench-med       & \textbf{70.00} & 65.00 & & 50.00 & \textbf{52.50} \\
\rowcolor{gray!12}
\quad \textbf{Avg. Performance}      & \textbf{79.33} & 68.44 & & 67.88 & 67.88 \\
\midrule
\multicolumn{6}{l}{\textit{Chat Hard Category}} \\
\quad mt-bench-hard         & \textbf{51.35} & 40.54 & & 54.05 & \textbf{59.46} \\
\quad llmbar-natural        & 46.00 & \textbf{55.00} & & 46.00 & 46.00 \\
\quad llmbar-adver-neighbor & 14.93 & \textbf{39.55} & & \textbf{44.78} & 35.07 \\
\quad llmbar-adver-GPTInst  & 16.30 & \textbf{22.83} & & \textbf{42.39} & 30.43 \\
\quad llmbar-adver-GPTOut   & 48.94 & 48.94 & & \textbf{48.94} & 46.81 \\
\quad llmbar-adver-manual   & 30.43 & \textbf{41.30} & & \textbf{32.61} & 26.09 \\
\rowcolor{gray!12}
\quad \textbf{Avg. Performance}     & 30.04 & \textbf{38.82} & & \textbf{44.52} & 38.82 \\
\midrule
\multicolumn{6}{l}{\textit{Reasoning Category}} \\
\quad hep-cpp    & 42.07 & \textbf{42.68} & & 43.90 & \textbf{51.22} \\
\quad hep-go     & 52.44 & 52.44 & & \textbf{45.12} & 39.02 \\
\quad hep-java   & \textbf{59.15} & 52.44 & & \textbf{50.61} & 47.56 \\
\quad hep-js     & \textbf{49.39} & 39.63 & & \textbf{55.49} & 49.39 \\
\quad hep-python & \textbf{57.93} & 50.00 & & 50.61 & 50.61 \\
\quad hep-rust   & \textbf{51.22} & 43.29 & & 43.90 & \textbf{45.73} \\
\quad math-prm   & \textbf{69.13} & 64.65 & & \textbf{97.54} & 82.99 \\
\rowcolor{gray!12}
\quad \textbf{Avg. Performance}  & \textbf{60.58} & 55.70 & & \textbf{72.91} & 65.13 \\
\midrule
\multicolumn{6}{l}{\textit{Safety Category}} \\
\quad refusals-dangerous    & \textbf{52.00} & 10.00 & & \textbf{50.00} & 45.00 \\
\quad refusals-offensive    & 47.00 & \textbf{63.00} & & \textbf{92.00} & 83.00 \\
\quad xstest-should-refuse  & \textbf{42.21} & 28.57 & & 60.39 & 60.39 \\
\quad xstest-should-respond & 61.60 & \textbf{72.40} & & \textbf{64.40} & 64.00 \\
\quad donotanswer           & \textbf{56.62} & 30.88 & & \textbf{40.44} & 36.76 \\
\rowcolor{gray!12}
\quad \textbf{Avg. Performance}     & \textbf{53.38} & 45.95 & & \textbf{60.95} & 58.24 \\
\midrule
\rowcolor{pajamablue}
\textbf{Overall} & \textbf{55.83} & 52.23 & & \textbf{61.57} & 57.52 \\
\bottomrule
\end{tabular}
\end{table}

\section{Ablation Studies}
\label{app:ablation_rubrics}

This section presents extensive ablation studies that isolate the contribution of each design choice in \textsc{Pajama}.
We first examine the two components of program synthesis: \textbf{(i)} the curated rubrics that steer program generation (Appendix~\ref{subsec:ablation_rubrics}) and \textbf{(ii)} the LLM that synthesizes programmatic judges (Appendix~\ref{subsec:ablation_synthesizer}).
We then examine two properties of the resulting committee: \textbf{(iii)} the number of programs aggregated into a verdict (Appendix~\ref{subsec:ablation_pool}), \textbf{(iv)} the number of labeled validation samples used to fit the aggregator (Appendix~\ref{subsec:ablation_val}), and \textbf{(v)} programs' transferability in a cross-dataset setting (Appendix~\ref{subsec:cross_dataset}).

\subsection{Synthesis Without Rubrics}
\label{subsec:ablation_rubrics}
\noindent\textbf{Setup.}
We first study the effect of evaluation rubrics on the synthesis of programmatic judges.
We synthesize an additional 80 programs with the rubric slot removed from the prompt, and then apply the same diversity filtering, selection, and aggregation pipeline to obtain final verdicts.
Any difference in performance is therefore attributable to the rubrics alone.

\noindent\textbf{Results.}
Without rubrics to steer synthesis, the generated programs collapse onto redundant logic and cover fewer useful judging dimensions: only 6 $\sim$ 8 programs survive filtering and selection, compared with 8 $\sim$ 21 under rubric guidance.
As Table~\ref{tab:ablation_naive} shows, this smaller committee yields lower test accuracy on four of the five datasets, with the largest gap on JudgeLM ($14.5$ points).
Rubrics are therefore a lightweight way to convey what the judging task requires, and the quality of the resulting committee depends on how well the synthesizer understands that requirement---a dependence we examine directly next.

\begin{table}[t!]
    \centering
    \small
    \caption{\textbf{Effect of removing rubric guidance from synthesis.} We report the number of programs surviving diversity filtering and selection, and the resulting test accuracy. Rubric guidance retains a larger, more diverse committee and improves accuracy on four of five datasets.}
    \label{tab:ablation_naive}
    \begin{tabular}{lcccc}
        \toprule
        & \multicolumn{2}{c}{\textbf{Na\"{i}ve Prompting}} & \multicolumn{2}{c}{\textbf{Rubric-Guided Prompting}} \\
        \cmidrule(lr){2-3} \cmidrule(lr){4-5}
        \textbf{Dataset} & \# Programs & Acc.\ (\%) & \# Programs & Acc.\ (\%) \\
        \midrule
        \textbf{JudgeLM}         & 8 & 66.62 & 21 & \textbf{81.13} \\
        \textbf{PandaLM}         & 6 & \textbf{71.49} & 14 & 70.38 \\
        \textbf{MultiPref}       & 6 & 67.09 & 16 & \textbf{72.23} \\
        \textbf{Prometheus}      & 6 & 82.09 &  8 & \textbf{88.78} \\
        \textbf{Preference-700K} & 6 & 67.71 & 15 & \textbf{67.82} \\
        \bottomrule
    \end{tabular}
\end{table}

\begin{table}[t!]
    \centering
    \small
    \caption{\textbf{Effect of the program synthesizer.} Test accuracy (\%) of programmatic judges synthesized by different LLMs. Stronger synthesizers produce stronger judges.}
    \label{tab:ablation_synthesizer}
    \begin{tabular}{lcc}
        \toprule
        \textbf{Program Synthesizer} & \textbf{Prometheus} & \textbf{JudgeLM} \\
        \midrule
        \textsc{GPT-4o}                      & 82.20 & 72.82 \\
        \textsc{GPT-5.5}                     & 82.46 & 80.01 \\
        \textsc{Claude Opus 4.6}$^{\dagger}$ & \textbf{88.78} & \textbf{81.13} \\
        \bottomrule
    \end{tabular}

    \vspace{2pt}
    {\footnotesize $^{\dagger}$Synthesizer used for all other experiments in this paper.}
\end{table}

\subsection{Choice of Program Synthesizer}
\label{subsec:ablation_synthesizer}
\noindent\textbf{Setup.}
The capability of the program synthesizer affects the quality of the generated programs, a point we also raise in our limitations (Appendix~\ref{app:discussion}).
To quantify this dependence, we synthesize programs with two additional LLMs (\textsc{GPT-4o} and \textsc{GPT-5.5}), holding all other components of the pipeline fixed, and compare the resulting programmatic judges on \textsc{Prometheus} and \textsc{JudgeLM}.

\noindent\textbf{Results.}
As Table~\ref{tab:ablation_synthesizer} shows, a weaker synthesizer degrades downstream evaluation accuracy: programs written by \textsc{GPT-4o} lose $6.6$ points on \textsc{Prometheus} and $8.3$ points on \textsc{JudgeLM} relative to those written by \textsc{Claude Opus 4.6}, the synthesizer used throughout the paper.
\textsc{GPT-5.5} recovers most of the gap on \textsc{JudgeLM} ($80.0$ vs.\ $81.1$) but remains $6.3$ points behind on \textsc{Prometheus}.
Taken together with the rubric ablation, this points to further headroom: \emph{stronger coding LLMs, or synthesis augmented with retrieval, expert domain knowledge, or more detailed task descriptions} can all be incorporated into \textsc{Pajama} to yield better programmatic judges\footnote{Given the rapid pace of progress in the coding capabilities of LLMs, we expect this limitation to diminish over time, with \textsc{Pajama} improving as stronger synthesizers become available.}.

\subsection{Effect of Program Pool Size}
\label{subsec:ablation_pool}
\noindent\textbf{Setup.}
We next study the effect of the committee size.
Intuitively, a larger committee offers more diverse judging perspectives for aggregation, but it also requires more program executions and higher synthesis cost, although both remain small in absolute terms.
To quantify this trade-off, we randomly sample $k$ programs from the synthesized pool and aggregate them into final verdicts, bypassing diversity filtering and selection so that $k$ is the only variable.
We run five trials for each value of $k$ and report accuracy averaged over the five datasets.

\noindent\textbf{Results.}
Table~\ref{tab:ablation_pool} shows that smaller committees cover less of the judging logic, exhibit higher variance across trials, and yield correspondingly lower test accuracy.
Returns diminish beyond roughly $k = 20$, where accuracy gains flatten out while the standard deviation continues to shrink.
Thus, a moderately sized committee of about 20 programs captures most of the benefit, and further additions mainly improve stability at a small extra cost.

\begin{table}[t!]
    \centering
    \small
    \caption{\textbf{Effect of program pool size.} Test accuracy (mean $\pm$ std over five trials, averaged across the five datasets) as a function of the number of randomly sampled programs $k$.}
    \label{tab:ablation_pool}
    \begin{tabular}{lcccccccc}
        \toprule
        \textbf{\# Programs ($k$)} & 1 & 3 & 5 & 10 & 20 & 30 & 40 & 50 \\
        \midrule
        \textbf{Mean Acc.\ (\%)}
          & 62.09 & 67.78 & 68.50 & 70.14 & 71.13 & 71.15 & 71.31 & \textbf{71.36} \\
        \textbf{Std}
          & 4.60 & 2.55 & 2.43 & 2.00 & 1.10 & 0.64 & 0.59 & 0.56 \\
        \bottomrule
    \end{tabular}
\end{table}

\begin{table}[t!]
\centering
\caption{
Testing accuracy (\%) as the size of the validation set is reduced.
For each ratio, we sample rows of the label matrix built from the validation set under 5 random seeds and report mean $\pm$ std.
}
\small
\setlength{\tabcolsep}{6pt}
\resizebox{\textwidth}{!}{
\begin{tabular}{lccccccc}
\toprule
& & \multicolumn{6}{c}{Fraction of validation set used} \\
\cmidrule(lr){3-8}
Dataset & $n_{\text{full}}$ & 100\% & 80\% & 60\% & 40\% & 20\% & 5\% \\
\midrule
\textbf{JudgeLM}          & 500 & $80.7 \pm 0.0$ & $80.8 \pm 0.2$ & $80.9 \pm 0.3$ & $80.6 \pm 0.6$ & $80.7 \pm 0.6$ & $80.5 \pm 0.6$ \\
\textbf{PandaLM}          & 500 & $69.8 \pm 0.0$ & $69.6 \pm 0.2$ & $69.4 \pm 0.2$ & $69.5 \pm 0.4$ & $69.2 \pm 0.3$ & $69.2 \pm 0.4$ \\
\textbf{MultiPref}        & 170 & $71.6 \pm 0.0$ & $72.0 \pm 0.2$ & $71.8 \pm 0.3$ & $72.0 \pm 0.2$ & $71.8 \pm 0.1$ & $71.4 \pm 1.3$ \\
\textbf{Prometheus}       & 500 & $88.0 \pm 0.0$ & $88.0 \pm 0.1$ & $88.0 \pm 0.1$ & $87.9 \pm 0.0$ & $88.2 \pm 0.0$ & $88.2 \pm 0.1$ \\
\textbf{Preference-700K}  & 500 & $67.4 \pm 0.0$ & $67.4 \pm 0.1$ & $67.5 \pm 0.1$ & $67.5 \pm 0.2$ & $67.5 \pm 0.1$ & $67.4 \pm 0.1$ \\
\bottomrule
\end{tabular}}
\label{tab:ablation_val_size}
\end{table}

\subsection{Effect of Validation Set Size}
\label{subsec:ablation_val}
\noindent\textbf{Setup.}
We further examine how the size of the validation set affects PAJAMA and how robust our evaluation performance is to this choice.
We vary the number of labeled samples $N$ used to fit the aggregator while holding the program committee fixed, and report the resulting test accuracy.

\noindent\textbf{Results.}
Table~\ref{tab:ablation_val_size} shows that test accuracy remains essentially flat as we shrink the validation set from $100\%$ to $5\%$: across all five datasets, performance moves by less than half a point in absolute terms, with no consistent trend in either direction.
This demonstrates our sample efficiency on providing reliable verdicts.
\textbf{\emph{The labeled-validation budget is therefore not a bottleneck for \textsc{Pajama}---a few dozen samples can learn an effective aggregator.}}

\begin{figure*}[t!]
    \centering
    \includegraphics[width=\textwidth]{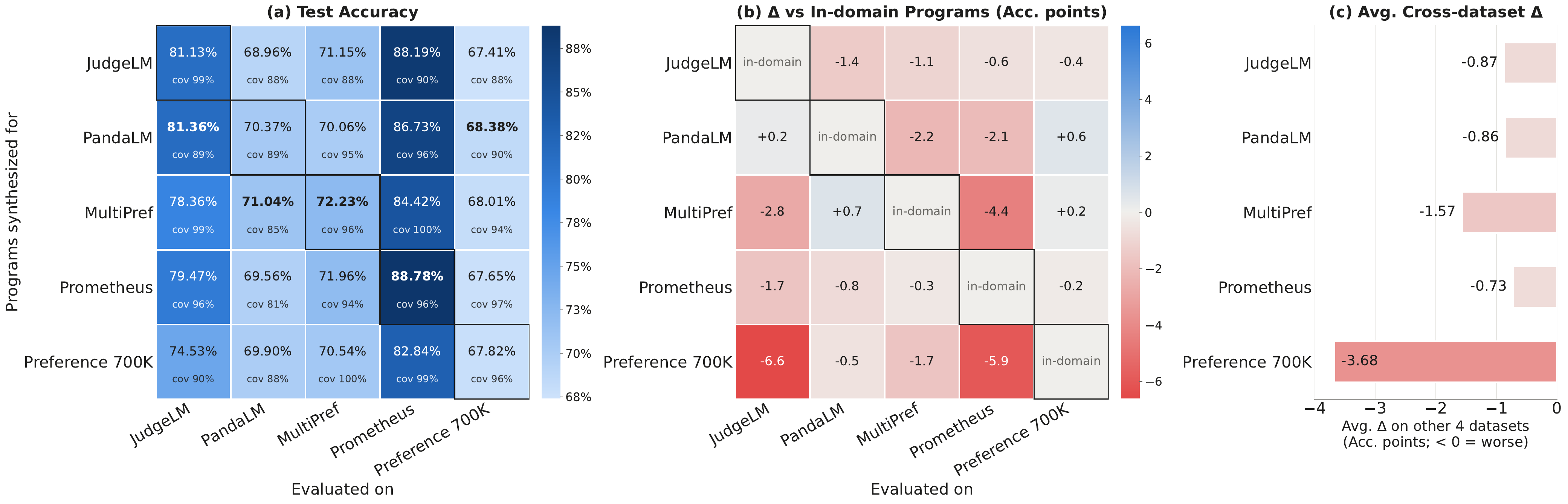}
    \caption{\textbf{Cross-dataset Transferability.}
    Rows: dataset the programs were synthesized for; columns: targeted evaluation dataset.
    (a) Test accuracy; outlined cells are in-domain.
    (b) Accuracy change relative to in-domain programs.
    (c) Average change when transferred to the other four datasets.}
    \label{fig:cross_dataset}
\end{figure*}

\subsection{Cross Dataset Transferability}
\label{subsec:cross_dataset}
\noindent\textbf{Setup.}
Finally, we test whether synthesized programs generalize beyond the dataset they were synthesized for, which indicates \textbf{\emph{whether a single, unified programmatic judge is feasible}}.
We apply the programs synthesized for each dataset to every other dataset, without re-synthesis.
Only our modeling steps (per-program thresholds, accuracy filtering, and top-$k$ selection) are refit on the target dataset's train and validation splits.

\noindent\textbf{Results.}
Figure~\ref{fig:cross_dataset} shows that programs transfer well, using in-domain programs (the diagonal) as the reference.
Across the 20 cross-dataset pairs, swapping in another dataset's programs costs only 1.6 accuracy points on average, and 15 of 20 pairs stay within 2 points of in-domain.
In 4 pairs, transferred programs even match or beat in-domain ones, e.g., Prometheus programs on MultiPref (+0.4).
Programs from Prometheus and PandaLM generalize best, losing less than 1 point on average.
Only when using Preference-700K, whose programs lose 3.68 points on average, mostly on JudgeLM ($-6.9$) and Prometheus ($-5.2$), suggesting its domain shift and encoding more dataset-specific criteria.
\emph{\textbf{Overall, synthesized programs capture largely reusable judging criteria: a program library built once can serve new datasets with only light recalibration.}}

\section{Discussion}
\label{app:discussion}

\noindent \textbf{Broader Impact.}
We do not foresee negative societal impacts from \textsc{PAJAMA}.
However, synthesized programs may inherit biases from an LLM when producing them, potentially leading to incorrect evaluations.
But this risk is unusually easy to detect and handle.

We can leverage several properties of programmatic judges that make the detection tractable.
\textbf{First}, programs expose fully transparent decision logic: expert users can read the code directly and verify which properties of a response actually drive the verdict.
\textbf{Second}, inspection need not be manual---modern coding agents can serve as automated diagnostic tools over the generated programs, as we demonstrate in \S\ref{subsec:bias_reduction}, where a second round of program calibration refines the committee and improves robustness on biased samples.
\textbf{Third}, a small validation set suffices as a probing dataset for analyzing program behavior---for instance, computing each program's coverage, conflict rate, and accuracy.
These diagnostics are simple, straightforward to run and provide concrete insight into program reliability.
\emph{\textbf{By contrast, comparable tools are largely unavailable for model-based evaluation, whose decision process remains opaque.}}

\noindent \textbf{Limitation.}
We identify potential limitations of \textsc{Pajama}, along with concrete ways to address them.

\textbf{First}, performance depends on the underlying LLM capabilities: if the model struggles to comprehend the task or generate effective programs, labeling quality degrades accordingly (see Appendix~\ref{subsec:ablation_synthesizer}).
This limitation, however, can be addressed with retrieval, with domain knowledge from subject-matter experts, or with more detailed task descriptions, and any advance in program synthesis can be incorporated into \textsc{Pajama} to yield better judges.

\textbf{Second}, \textsc{Pajama} is best suited to responses that admit relatively direct evaluation; on complex reasoning tasks, synthesized programs may fail to generalize.
We study two mitigations, both validated empirically.
One is to route low-confidence samples to an LLM judge that can handle the harder cases: with programmatic judges acting as an efficient first-pass filter, this hybrid design pushes the accuracy--throughput frontier outward (\S\ref{subsec:pareto_frontier}).
The other is to distill program verdicts into a reward model, which generalizes more strongly than the programs themselves and outperforms frontier models on reasoning and safety categories at $50\times$ lower cost (\S\ref{subsec:reward_model}).

\end{document}